\documentclass[twoside,11pt]{article}

\usepackage{jmlr2e} 
\usepackage{amsmath,epsfig} 
\usepackage{amsfonts}
\usepackage{xcolor}
\usepackage{amssymb}
\usepackage{subfigure}
\usepackage[algo2e]{algorithm2e}
\usepackage{paralist}
\usepackage{booktabs}
\usepackage{subfigure}
\usepackage{color}
\DeclareGraphicsExtensions{.pdf,.jpeg,.png,.epbibdesks}
\usepackage{epstopdf}
\usepackage[T1]{fontenc}
\usepackage{graphicx}
\usepackage{algpseudocode}
\usepackage{url}
\usepackage{multicol}

\usepackage{amssymb}
\usepackage{amsfonts}
\usepackage{mathrsfs}
\usepackage{xspace}
\usepackage{bm}
\usepackage{upgreek}

\newcommand{\safemath}[2]{\newcommand{#1}{\ensuremath{#2}\xspace}}

\safemath{\bma}{\mathbf{a}}
\safemath{\bmb}{\mathbf{b}}
\safemath{\bmc}{\mathbf{c}}
\safemath{\bmd}{\mathbf{d}}
\safemath{\bme}{\mathbf{e}}
\safemath{\bmf}{\mathbf{f}}
\safemath{\bmg}{\mathbf{g}}
\safemath{\bmh}{\mathbf{h}}
\safemath{\bmi}{\mathbf{i}}
\safemath{\bmj}{\mathbf{j}}
\safemath{\bmk}{\mathbf{k}}
\safemath{\bml}{\mathbf{l}}
\safemath{\bmm}{\mathbf{m}}
\safemath{\bmn}{\mathbf{n}}
\safemath{\bmo}{\mathbf{o}}
\safemath{\bmp}{\mathbf{p}}
\safemath{\bmq}{\mathbf{q}}
\safemath{\bmr}{\mathbf{r}}
\safemath{\bms}{\mathbf{s}}
\safemath{\bmt}{\mathbf{t}}
\safemath{\bmu}{\mathbf{u}}
\safemath{\bmv}{\mathbf{v}}
\safemath{\bmw}{\mathbf{w}}
\safemath{\bmx}{\mathbf{x}}
\safemath{\bmy}{\mathbf{y}}
\safemath{\bmz}{\mathbf{z}}
\safemath{\bmzero}{\mathbf{0}}
\safemath{\bmone}{\mathbf{1}}

\bmdefine{\biad}{a}
\bmdefine{\bibd}{b}
\bmdefine{\bicd}{c}
\bmdefine{\bidd}{d}
\bmdefine{\bied}{e}
\bmdefine{\bifd}{f}
\bmdefine{\bigd}{g}
\bmdefine{\bihd}{h}
\bmdefine{\biid}{i}
\bmdefine{\bijd}{j}
\bmdefine{\bikd}{k}
\bmdefine{\bild}{l}
\bmdefine{\bimd}{m}
\bmdefine{\bind}{n}
\bmdefine{\biod}{o}
\bmdefine{\bipd}{p}
\bmdefine{\biqd}{q}
\bmdefine{\bird}{r}
\bmdefine{\bisd}{s}
\bmdefine{\bitd}{t}
\bmdefine{\biud}{u}
\bmdefine{\bivd}{v}
\bmdefine{\biwd}{w}
\bmdefine{\bixd}{x}
\bmdefine{\biyd}{y}
\bmdefine{\bizd}{z}

\bmdefine{\bixid}{\xi}
\bmdefine{\bilambdad}{\lambda}
\bmdefine{\bimud}{\mu}
\bmdefine{\bithetad}{\theta}
\bmdefine{\biphid}{\phi}
\bmdefine{\bideltad}{\delta}

\safemath{\bmia}{\biad}
\safemath{\bmib}{\bibd}
\safemath{\bmic}{\bicd}
\safemath{\bmid}{\bidd}
\safemath{\bmie}{\bied}
\safemath{\bmif}{\bifd}
\safemath{\bmig}{\bigd}
\safemath{\bmih}{\bihd}
\safemath{\bmii}{\biid}
\safemath{\bmij}{\bijd}
\safemath{\bmik}{\bikd}
\safemath{\bmil}{\bild}
\safemath{\bmim}{\bimd}
\safemath{\bmin}{\bind}
\safemath{\bmio}{\biod}
\safemath{\bmip}{\bipd}
\safemath{\bmiq}{\biqd}
\safemath{\bmir}{\bird}
\safemath{\bmis}{\bisd}
\safemath{\bmit}{\bitd}
\safemath{\bmiu}{\biud}
\safemath{\bmiv}{\bivd}
\safemath{\bmiw}{\biwd}
\safemath{\bmix}{\bixd}
\safemath{\bmiy}{\biyd}
\safemath{\bmiz}{\bizd}

\safemath{\bmxi}{\bixid}
\safemath{\bmlambda}{\bilambdad}
\safemath{\bmmu}{\bimud}
\safemath{\bmtheta}{\bithetad}
\safemath{\bmphi}{\biphid}
\safemath{\bmdelta}{\bideltad}

\safemath{\bA}{\mathbf{A}}
\safemath{\bB}{\mathbf{B}}
\safemath{\bC}{\mathbf{C}}
\safemath{\bD}{\mathbf{D}}
\safemath{\bE}{\mathbf{E}}
\safemath{\bF}{\mathbf{F}}
\safemath{\bG}{\mathbf{G}}
\safemath{\bH}{\mathbf{H}}
\safemath{\bI}{\mathbf{I}}
\safemath{\bJ}{\mathbf{J}}
\safemath{\bK}{\mathbf{K}}
\safemath{\bL}{\mathbf{L}}
\safemath{\bM}{\mathbf{M}}
\safemath{\bN}{\mathbf{N}}
\safemath{\bO}{\mathbf{O}}
\safemath{\bP}{\mathbf{P}}
\safemath{\bQ}{\mathbf{Q}}
\safemath{\bR}{\mathbf{R}}
\safemath{\bS}{\mathbf{S}}
\safemath{\bT}{\mathbf{T}}
\safemath{\bU}{\mathbf{U}}
\safemath{\bV}{\mathbf{V}}
\safemath{\bW}{\mathbf{W}}
\safemath{\bX}{\mathbf{X}}
\safemath{\bY}{\mathbf{Y}}
\safemath{\bZ}{\mathbf{Z}}

\safemath{\bZero}{\mathbf{0}}
\safemath{\bOne}{\mathbf{1}}
\safemath{\bDelta}{\mathbf{\Delta}}
\safemath{\bLambda}{\mathbf{\UpLambda}}
\safemath{\bPhi}{\mathbf{\Upphi}}
\safemath{\bSigma}{\mathbf{\Upsigma}}
\safemath{\bOmega}{\mathbf{\Upomega}}
\safemath{\bTheta}{\mathbf{\Uptheta}}

\bmdefine{\biAd}{A}
\bmdefine{\biBd}{B}
\bmdefine{\biCd}{C}
\bmdefine{\biDd}{D}
\bmdefine{\biEd}{E}
\bmdefine{\biFd}{F}
\bmdefine{\biGd}{G}
\bmdefine{\biHd}{H}
\bmdefine{\biId}{I}
\bmdefine{\biJd}{J}
\bmdefine{\biKd}{K}
\bmdefine{\biLd}{L}
\bmdefine{\biMd}{M}
\bmdefine{\biNd}{N}
\bmdefine{\biOd}{O}
\bmdefine{\biPd}{P}
\bmdefine{\biQd}{Q}
\bmdefine{\biRd}{R}
\bmdefine{\biSd}{S}
\bmdefine{\biTd}{T}
\bmdefine{\biUd}{U}
\bmdefine{\biVd}{V}
\bmdefine{\biWd}{W}
\bmdefine{\biXd}{X}
\bmdefine{\biYd}{Y}
\bmdefine{\biZd}{Z}

\bmdefine{\biDelta}{\Delta}
\bmdefine{\biLambda}{\Lambda}
\bmdefine{\biPhi}{\Phi}
\bmdefine{\biSigma}{\Sigma}
\bmdefine{\biOmega}{\Omega}
\bmdefine{\biTheta}{\Theta}

\safemath{\bimA}{\biAd}
\safemath{\bimB}{\biBd}
\safemath{\bimC}{\biCd}
\safemath{\bimD}{\biDd}
\safemath{\bimE}{\biEd}
\safemath{\bimF}{\biFd}
\safemath{\bimG}{\biGd}
\safemath{\bimH}{\biHd}
\safemath{\bimI}{\biId}
\safemath{\bimJ}{\biJd}
\safemath{\bimK}{\biKd}
\safemath{\bimL}{\biLd}
\safemath{\bimM}{\biMd}
\safemath{\bimN}{\biNd}
\safemath{\bimO}{\biOd}
\safemath{\bimP}{\biPd}
\safemath{\bimQ}{\biQd}
\safemath{\bimR}{\biRd}
\safemath{\bimS}{\biSd}
\safemath{\bimT}{\biTd}
\safemath{\bimU}{\biUd}
\safemath{\bimV}{\biVd}
\safemath{\bimW}{\biWd}
\safemath{\bimX}{\biXd}
\safemath{\bimY}{\biYd}
\safemath{\bimZ}{\biZd}

\safemath{\bimDelta}{\biDelta}
\safemath{\bimLambda}{\biLambda}
\safemath{\bimPhi}{\biPhi}
\safemath{\bimSigma}{\biSigma}
\safemath{\bimOmega}{\biOmega}
\safemath{\bimTheta}{\biTheta}

\safemath{\setA}{\mathcal{A}}
\safemath{\setB}{\mathcal{B}}
\safemath{\setC}{\mathcal{C}}
\safemath{\setD}{\mathcal{D}}
\safemath{\setE}{\mathcal{E}}
\safemath{\setF}{\mathcal{F}}
\safemath{\setG}{\mathcal{G}}
\safemath{\setH}{\mathcal{H}}
\safemath{\setI}{\mathcal{I}}
\safemath{\setJ}{\mathcal{J}}
\safemath{\setK}{\mathcal{K}}
\safemath{\setL}{\mathcal{L}}
\safemath{\setM}{\mathcal{M}}
\safemath{\setN}{\mathcal{N}}
\safemath{\setO}{\mathcal{O}}
\safemath{\setP}{\mathcal{P}}
\safemath{\setQ}{\mathcal{Q}}
\safemath{\setR}{\mathcal{R}}
\safemath{\setS}{\mathcal{S}}
\safemath{\setT}{\mathcal{T}}
\safemath{\setU}{\mathcal{U}}
\safemath{\setV}{\mathcal{V}}
\safemath{\setW}{\mathcal{W}}
\safemath{\setX}{\mathcal{X}}
\safemath{\setY}{\mathcal{Y}}
\safemath{\setZ}{\mathcal{Z}}
\safemath{\emptySet}{\varnothing}

\safemath{\colA}{\mathscr{A}}
\safemath{\colB}{\mathscr{B}}
\safemath{\colC}{\mathscr{C}}
\safemath{\colD}{\mathscr{D}}
\safemath{\colE}{\mathscr{E}}
\safemath{\colF}{\mathscr{F}}
\safemath{\colG}{\mathscr{G}}
\safemath{\colH}{\mathscr{H}}
\safemath{\colI}{\mathscr{I}}
\safemath{\colJ}{\mathscr{J}}
\safemath{\colK}{\mathscr{K}}
\safemath{\colL}{\mathscr{L}}
\safemath{\colM}{\mathscr{M}}
\safemath{\colN}{\mathscr{N}}
\safemath{\colO}{\mathscr{O}}
\safemath{\colP}{\mathscr{P}}
\safemath{\colQ}{\mathscr{Q}}
\safemath{\colR}{\mathscr{R}}
\safemath{\colS}{\mathscr{S}}
\safemath{\colT}{\mathscr{T}}
\safemath{\colU}{\mathscr{U}}
\safemath{\colV}{\mathscr{V}}
\safemath{\colW}{\mathscr{W}}
\safemath{\colX}{\mathscr{X}}
\safemath{\colY}{\mathscr{Y}}
\safemath{\colZ}{\mathscr{Z}}

\safemath{\opA}{\mathbb{A}}
\safemath{\opB}{\mathbb{B}}
\safemath{\opC}{\mathbb{C}}
\safemath{\opD}{\mathbb{D}}
\safemath{\opE}{\mathbb{E}}
\safemath{\opF}{\mathbb{F}}
\safemath{\opG}{\mathbb{G}}
\safemath{\opH}{\mathbb{H}}
\safemath{\opI}{\mathbb{I}}
\safemath{\opJ}{\mathbb{J}}
\safemath{\opK}{\mathbb{K}}
\safemath{\opL}{\mathbb{L}}
\safemath{\opM}{\mathbb{M}}
\safemath{\opN}{\mathbb{N}}
\safemath{\opO}{\mathbb{O}}
\safemath{\opP}{\mathbb{P}}
\safemath{\opQ}{\mathbb{Q}}
\safemath{\opR}{\mathbb{R}}
\safemath{\opS}{\mathbb{S}}
\safemath{\opT}{\mathbb{T}}
\safemath{\opU}{\mathbb{U}}
\safemath{\opV}{\mathbb{V}}
\safemath{\opW}{\mathbb{W}}
\safemath{\opX}{\mathbb{X}}
\safemath{\opY}{\mathbb{Y}}
\safemath{\opZ}{\mathbb{Z}}
\safemath{\opZero}{\mathbb{O}}
\safemath{\identityop}{\opI}

\safemath{\veca}{\bma}
\safemath{\vecb}{\bmb}
\safemath{\vecc}{\bmc}
\safemath{\vecd}{\bmd}
\safemath{\vece}{\bme}
\safemath{\vecf}{\bmf}
\safemath{\vecg}{\bmg}
\safemath{\vech}{\bmh}
\safemath{\veci}{\bmi}
\safemath{\vecj}{\bmj}
\safemath{\veck}{\bmk}
\safemath{\vecl}{\bml}
\safemath{\vecm}{\bmm}
\safemath{\vecn}{\bmn}
\safemath{\veco}{\bmo}
\safemath{\vecp}{\bmp}
\safemath{\vecq}{\bmq}
\safemath{\vecr}{\bmr}
\safemath{\vecs}{\bms}
\safemath{\vect}{\bmt}
\safemath{\vecu}{\bmu}
\safemath{\vecv}{\bmv}
\safemath{\vecw}{\bmw}
\safemath{\vecx}{\bmx}
\safemath{\vecy}{\bmy}
\safemath{\vecz}{\bmz}

\safemath{\veczero}{\bmzero}
\safemath{\vecone}{\bmone}
\safemath{\vecxi}{\bmxi}
\safemath{\veclambda}{\bmlambda}
\safemath{\vecmu}{\bmmu}
\safemath{\vectheta}{\bmtheta}
\safemath{\vecphi}{\bmphi}
\safemath{\vecdelta}{\bmdelta}

\safemath{\matA}{\bA}
\safemath{\matB}{\bB}
\safemath{\matC}{\bC}
\safemath{\matD}{\bD}
\safemath{\matE}{\bE}
\safemath{\matF}{\bF}
\safemath{\matG}{\bG}
\safemath{\matH}{\bH}
\safemath{\matI}{\bI}
\safemath{\matJ}{\bJ}
\safemath{\matK}{\bK}
\safemath{\matL}{\bL}
\safemath{\matM}{\bM}
\safemath{\matN}{\bN}
\safemath{\matO}{\bO}
\safemath{\matP}{\bP}
\safemath{\matQ}{\bQ}
\safemath{\matR}{\bR}
\safemath{\matS}{\bS}
\safemath{\matT}{\bT}
\safemath{\matU}{\bU}
\safemath{\matV}{\bV}
\safemath{\matW}{\bW}
\safemath{\matX}{\bX}
\safemath{\matY}{\bY}
\safemath{\matZ}{\bZ}
\safemath{\matzero}{\bmzero}

\safemath{\matDelta}{\bDelta}
\safemath{\matLambda}{\bLambda}
\safemath{\matPhi}{\bPhi}
\safemath{\matSigma}{\bSigma}
\safemath{\matOmega}{\bOmega}
\safemath{\matTheta}{\bTheta}

\safemath{\matidentity}{\matI}
\safemath{\matone}{\matO}

\safemath{\rnda}{A}
\safemath{\rndb}{B}
\safemath{\rndc}{C}
\safemath{\rndd}{D}
\safemath{\rnde}{E}
\safemath{\rndf}{F}
\safemath{\rndg}{G}
\safemath{\rndh}{H}
\safemath{\rndi}{I}
\safemath{\rndj}{J}
\safemath{\rndk}{K}
\safemath{\rndl}{L}
\safemath{\rndm}{M}
\safemath{\rndn}{N}
\safemath{\rndo}{O}
\safemath{\rndp}{P}
\safemath{\rndq}{Q}
\safemath{\rndr}{R}
\safemath{\rnds}{S}
\safemath{\rndt}{T}
\safemath{\rndu}{U}
\safemath{\rndv}{V}
\safemath{\rndw}{W}
\safemath{\rndx}{X}
\safemath{\rndy}{Y}
\safemath{\rndz}{Z}

\safemath{\rveca}{\bimA}
\safemath{\rvecb}{\bimB}
\safemath{\rvecc}{\bimC}
\safemath{\rvecd}{\bimD}
\safemath{\rvece}{\bimE}
\safemath{\rvecf}{\bimF}
\safemath{\rvecg}{\bimG}
\safemath{\rvech}{\bimH}
\safemath{\rveci}{\bimI}
\safemath{\rvecj}{\bimJ}
\safemath{\rveck}{\bimK}
\safemath{\rvecl}{\bimL}
\safemath{\rvecm}{\bimM}
\safemath{\rvecn}{\bimN}
\safemath{\rveco}{\bomO}
\safemath{\rvecp}{\bimP}
\safemath{\rvecq}{\bimQ}
\safemath{\rvecr}{\bimR}
\safemath{\rvecs}{\bimS}
\safemath{\rvect}{\bimT}
\safemath{\rvecu}{\bimU}
\safemath{\rvecv}{\bimV}
\safemath{\rvecw}{\bimW}
\safemath{\rvecx}{\bimX}
\safemath{\rvecy}{\bimY}
\safemath{\rvecz}{\bimZ}

\safemath{\rvecxi}{\bmxi}
\safemath{\rveclambda}{\bmlambda}
\safemath{\rvecmu}{\bmmu}
\safemath{\rvectheta}{\bmtheta}
\safemath{\rvecphi}{\bmphi}

\safemath{\rmatA}{\bimA}
\safemath{\rmatB}{\bimB}
\safemath{\rmatC}{\bimC}
\safemath{\rmatD}{\bimD}
\safemath{\rmatE}{\bimE}
\safemath{\rmatF}{\bimF}
\safemath{\rmatG}{\bimG}
\safemath{\rmatH}{\bimH}
\safemath{\rmatI}{\bimI}
\safemath{\rmatJ}{\bimJ}
\safemath{\rmatK}{\bimK}
\safemath{\rmatL}{\bimL}
\safemath{\rmatM}{\bimM}
\safemath{\rmatN}{\bimN}
\safemath{\rmatO}{\bimO}
\safemath{\rmatP}{\bimP}
\safemath{\rmatQ}{\bimQ}
\safemath{\rmatR}{\bimR}
\safemath{\rmatS}{\bimS}
\safemath{\rmatT}{\bimT}
\safemath{\rmatU}{\bimU}
\safemath{\rmatV}{\bimV}
\safemath{\rmatW}{\bimW}
\safemath{\rmatX}{\bimX}
\safemath{\rmatY}{\bimY}
\safemath{\rmatZ}{\bimZ}

\safemath{\rmatDelta}{\bimDelta}
\safemath{\rmatLambda}{\bimLambda}
\safemath{\rmatPhi}{\bimPhi}
\safemath{\rmatSigma}{\bimSigma}
\safemath{\rmatOmega}{\bimOmega}
\safemath{\rmatTheta}{\bimTheta}

\usepackage{amssymb}
\usepackage{amsfonts}
\usepackage{mathrsfs}
\usepackage{xspace}
\usepackage{bm}
\usepackage{fancyref}
\usepackage{textcomp}

\usepackage{multirow}
\usepackage{stmaryrd}

\newenvironment{textbmatrix}{	\setlength{\arraycolsep}{2.5pt}%
								\big[\begin{matrix}}{\end{matrix}\big]%
								\raisebox{0.08ex}{\vphantom{M}}}

\def\be{\begin{equation}}
\def\ee{\end{equation}}
\def\een{\nonumber \end{equation}}
\def\mat{\begin{bmatrix}}
\def\emat{\end{bmatrix}}
\def\btm{\begin{textbmatrix}}
\def\etm{\end{textbmatrix}}

\def\ba#1\ea{\begin{align}#1\end{align}}
\def\bas#1\eas{\begin{align*}#1\end{align*}}
\def\bs#1\es{\begin{split}#1\end{split}} 
\def\bg#1\eg{\begin{gather}#1\end{gather}}
\def\bml#1\eml{\begin{multline}#1\end{multline}}
\def\bi#1\ei{\begin{itemize}#1\end{itemize}}

\newcommand{\lefto}{\mathopen{}\left}

\newcommand{\vecnorm}[1]{\lefto\lVert#1\right\rVert}		

\safemath{\dirac}{\delta}					
\safemath{\krond}{\dirac}					

\safemath{\upto}{\uparrow}
\safemath{\downto}{\downarrow}
\safemath{\iu}{j}							
\safemath{\ev}{\lambda}						
\safemath{\hilseqspace}{l^{2}}				
\newcommand{\banachfunspace}[1]{\setL^{#1}}	
\safemath{\hilfunspace}{\banachfunspace{2}}	

\safemath{\SNR}{\text{\sc snr}} 				
\safemath{\No}{N_0}							
\safemath{\Es}{E_s}							
\safemath{\Eb}{E_b}							
\safemath{\EbNo}{\frac{\Eb}{\No}}
\safemath{\EsNo}{\frac{\Es}{\No}}

\DeclareMathOperator{\CHop}{\ensuremath{\opH}} 
\safemath{\tvir}{\rndh_{\CHop}}				
\safemath{\tvtf}{\rndl_{\CHop}}				
\safemath{\spf}{\rnds_{\CHop}}				
\safemath{\bff}{H_{\CHop}}					

\safemath{\ircf}{r_{h}}						
\safemath{\tftvcf}{r_{s}}					
\safemath{\tfcf}{r_{l}}						
\safemath{\bfcf}{r_{H}}						

\safemath{\tcorr}{c_h}						
\safemath{\scf}{c_{s}}						
\safemath{\tfcorr}{c_{l}}					
\safemath{\fcorr}{c_{H}}						

\safemath{\mi}{I}							
\safemath{\capacity}{C}						

\safemath{\normal}{\mathcal{N}}			
\safemath{\jpg}{\mathcal{CN}}			
\safemath{\mchain}{\leftrightarrow}		

\safemath{\dB}{\,\mathrm{dB}}
\safemath{\dBm}{\,\mathrm{dBm}}
\safemath{\Hz}{\,\mathrm{Hz}}
\safemath{\kHz}{\,\mathrm{kHz}}
\safemath{\MHz}{\,\mathrm{MHz}}
\safemath{\GHz}{\,\mathrm{GHz}}
\safemath{\s}{\,\mathrm{s}}
\safemath{\ms}{\,\mathrm{ms}}
\safemath{\mus}{\,\mathrm{\text{\textmu}s}}
\safemath{\ns}{\,\mathrm{ns}}
\safemath{\ps}{\,\mathrm{ps}}
\safemath{\meter}{\,\mathrm{m}}
\safemath{\mm}{\,\mathrm{mm}}
\safemath{\cm}{\,\mathrm{cm}}
\safemath{\m}{\,\mathrm{m}}
\safemath{\W}{\,\mathrm{W}}
\safemath{\mW}{\, \mathrm{mW}}
\safemath{\J}{\,\mathrm{J}}
\safemath{\K}{\,\mathrm{K}}
\safemath{\bit}{\,\mathrm{bit}}
\safemath{\nat}{\,\mathrm{nat}}

\safemath{\define}{\triangleq}			

\safemath{\equivalent}{\sim}
\safemath{\distas}{\sim}					
\safemath{\sdiff}{\Delta}				

\safemath{\reals}{\mathbb{R}}
\safemath{\positivereals}{\reals_{+}}
\safemath{\integers}{\mathbb{Z}}
\safemath{\posint}{\integers_{+}}
\safemath{\naturals}{\mathbb{N}}
\safemath{\posnaturals}{\naturals_{+}}
\safemath{\complexset}{\mathbb{C}}
\safemath{\rationals}{\mathbb{Q}}

\newcommand*{\fancyrefapplabelprefix}{app}		
\newcommand*{\fancyrefthmlabelprefix}{thm}		
\newcommand*{\fancyreflemlabelprefix}{lem}		
\newcommand*{\fancyrefcorlabelprefix}{cor}		
\newcommand*{\fancyrefdeflabelprefix}{def}		
\newcommand*{\fancyrefalglabelprefix}{alg}		
\newcommand*{\fancyrefproplabelprefix}{prop}		
\newcommand*{\fancyrefexmpllabelprefix}{exmpl}
\newcommand*{\fancyreftbllabelprefix}{tbl}
\frefformat{vario}{\fancyrefseclabelprefix}{Section~#1}
\frefformat{vario}{\fancyrefthmlabelprefix}{Theorem~#1}
\frefformat{vario}{\fancyreflemlabelprefix}{Lemma~#1}
\frefformat{vario}{\fancyrefcorlabelprefix}{Corrolary~#1}
\frefformat{vario}{\fancyrefdeflabelprefix}{Definition~#1}
\frefformat{vario}{\fancyrefalglabelprefix}{Algorithm~#1}
\frefformat{vario}{\fancyreffiglabelprefix}{Figure~#1}
\frefformat{vario}{\fancyrefapplabelprefix}{Appendix~#1} 
\frefformat{vario}{\fancyrefeqlabelprefix}{(#1)}
\frefformat{vario}{\fancyrefproplabelprefix}{Proposition~#1}
\frefformat{vario}{\fancyrefexmpllabelprefix}{Example~#1}
\frefformat{vario}{\fancyreftbllabelprefix}{Table~#1}

\safemath{\dictab}{[\,\dicta\,\,\dictb\,]}

\safemath{\ysig}{\bmy}
\safemath{\ysighat}{\hat{\ysig}}
\safemath{\ysigdim}{M}
\safemath{\xsig}{\bmx}
\safemath{\xsigdim}{N}
\safemath{\nx}{n_x}
\safemath{\zsig}{\bmz}
\safemath{\zsigdim}{\ysigdim}
\safemath{\rsig}{\bmr}
\safemath{\Adict}{\bA}
\safemath{\Adicttilde}{\widetilde{\Adict}}
\safemath{\Adictdim}{\outputdim\times\xsigdim}
\safemath{\avec}{\bma}
\safemath{\avectilde}{\tilde{\avec}}
\safemath{\Bdict}{\bB}
\safemath{\Bdicttilde}{\widetilde{\Bdict}}
\safemath{\Cdict}{\bC}
\safemath{\cvec}{\bmc}
\safemath{\Ddict}{\bD}
\safemath{\Ddictdim}{\ysigdim\times\xsigdim}
\safemath{\dvec}{\bmd}
\safemath{\Ddicttilde}{\widetilde{\bD}}
\safemath{\Bonb}{\bB}
\safemath{\bvec}{\bmb}
\safemath{\Bonbdim}{\ysigdim\times\ysigdim}
\safemath{\noise}{\bmn}
\safemath{\noisedim}{\ysigim}
\safemath{\err}{\bme}
\safemath{\errdim}{\ysigdim}
\safemath{\errset}{\setE}
\safemath{\nerr}{n_e}
\safemath{\delop}{\bP_\errset}
\safemath{\delopc}{\bP_{{\errset}^c}}

\safemath{\cplxi}{\imath}
\safemath{\cplxj}{\jmath}

\safemath{\dict}{\matD}
\safemath{\inputdim}{N}		
\safemath{\outputdim}{M}		
\safemath{\sparsity}{S}	
\safemath{\inputdimA}{{N_a}}	
\safemath{\inputdimB}{{N_b}}	
\safemath{\elemA}{{n_a}}	
\safemath{\elemB}{{n_b}}	
\safemath{\resA}{\matR_a}	
\safemath{\resB}{\matR_b}	
\safemath{\subD}{\matS} 
\safemath{\subA}{\matS_a} 
\safemath{\subB}{\matS_b} 
\safemath{\dicta}{\matA} 	
\safemath{\dictb}{\matB} 	
\safemath{\hollowS}{H}
\safemath{\hollowA}{H_a}
\safemath{\hollowB}{H_b}
\safemath{\cross}{Z}
\safemath{\coh}{\mu_d}			
\safemath{\coha}{\mu_a}			
\safemath{\cohb}{\mu_b}			
\safemath{\mubs}{\nu}	
\safemath{\cohm}{\mu_m} 
\safemath{\dictset}{\setD}	
\safemath{\dictsetp}{\dictset(\coh,\coha,\cohb)}	
\safemath{\dictsetgen}{\dictset_\text{gen}}
\safemath{\dictsetgenp}{\dictsetgen(\coh)}
\safemath{\dictsetonb}{\dictset_\text{onb}}
\safemath{\dictsetonbp}{\dictsetonb(\coh)}

\safemath{\leftside}{U}
\safemath{\rightsideA}{R_a}
\safemath{\rightsideB}{R_b}

\safemath{\indexS}{\setI_S} 

\safemath{\na}{n_a}			
\safemath{\nb}{n_b}			
\safemath{\coeffa}{p_i}	
\safemath{\coeffb}{q_j}	
\safemath{\seta}{\setP}		
\safemath{\setb}{\setQ}     
\safemath{\setw}{\setW}	
\safemath{\setz}{\setZ}	
\safemath{\cola}{\veca}		
\safemath{\colb}{\vecb}		
\safemath{\cold}{\vecd}		
\safemath{\inputvec}{\vecx} 	
\safemath{\error}{\vece}	
\safemath{\noiseout}{\vecz} 	
\safemath{\inputvecel}{x}
\safemath{\inputveca}{\vecx_a}
\safemath{\inputvecb}{\vecx_b}
\safemath{\outputvec}{\vecy}	
\safemath{\lambdamin}{\lambda_{\mathrm{min}}}

\newcommand{\normone}[1]{\vecnorm{#1}_1}

\safemath{\elltwo}{\ell_2}
\safemath{\ellone}{\ell_1}
\safemath{\ellzero}{\ell_0}
\safemath{\ellinf}{\ell_\infty}
\safemath{\licard}{Z(\coh,\coha,\cohb)}
\safemath{\xsol}{\hat{x}}
\safemath{\xbord}{x_b}		
\safemath{\xstat}{x_s}		
\safemath{\xstatLone}{\tilde{x}_s}
\safemath{\order}{\mathcal{O}} 
\safemath{\scales}{\Theta} 
\safemath{\ones}{\mathbf{1}} 
\safemath{\zeroes}{\mathbf{0}} 
\safemath{\thlone}{\kappa(\coh,\cohb)} 
\safemath{\constoneA}{\delta} 
\safemath{\constoneB}{\epsilon} 
\safemath{\nlarge}{L}				   
\safemath{\sumlarge}{S_\nlarge}
\safemath{\maxlarger}{P_\nlarge}	   
\safemath{\Pzero}{\textrm{P0}}	
\safemath{\Pone}{\textrm{P1}}
\safemath{\vecfir}{\vecw}			 
\safemath{\vecsec}{\vecz}
\safemath{\elvecfir}{w}              
\safemath{\elvecsec}{z}				 
\safemath{\nlargefir}{n}
\safemath{\normout}{\gamma}
\safemath{\auxfun}{h}
\safemath{\supp}{\textrm{supp}}

\safemath{\indexa}{\ell}
\safemath{\indexb}{r}
\safemath{\indexc}{i}
\safemath{\indexd}{j}

\safemath{\project}{P}
\renewcommand{\vecw}{\bar{\mathbf{w}}}

\firstpageno{1}

\begin{document}

\title{Time-varying Learning and Content Analytics via Sparse Factor Analysis}

\author{\name Andrew\ S.\ Lan \email mr.lan@sparfa.com \\
	\name Christoph Studer \email studer@sparfa.com \\
	\name Richard G. Baraniuk \email richb@sparfa.com \\
       \addr Dept.~Electrical and Computer Engineering\\
       Rice University\\
       Houston, TX 77005, USA}

   \editor{}

\maketitle

\begin{abstract}
We propose SPARFA-Trace, a new machine learning-based framework for time-varying \emph{learning} and \emph{content analytics} for education applications. 
We develop a novel message passing-based, blind, approximate Kalman filter for sparse factor analysis (SPARFA), that jointly (i) traces learner concept knowledge over time, (ii) analyzes learner concept knowledge state transitions (induced by interacting with learning resources, such as textbook sections, lecture videos, etc, or the forgetting effect), and (iii) estimates the content organization and intrinsic difficulty of the assessment questions. 
These quantities are estimated solely from binary-valued (correct/incorrect) graded learner response data and a summary of the specific actions each learner performs (e.g., answering a question or studying a learning resource) at each time instance. 
Experimental results on two online course datasets demonstrate that SPARFA-Trace is capable of tracing each learner's concept knowledge evolution over time, as well as analyzing the quality and content organization of learning resources, the question--concept associations, and the question intrinsic difficulties. 
Moreover, we show that SPARFA-Trace achieves comparable or better performance in predicting unobserved learner responses than existing collaborative filtering and knowledge tracing approaches for personalized education.
\end{abstract}


\section{Introduction}
\label{sec:intro}

The traditional ``one-size-fits-all" education approach is one of the main bottlenecks for education in the $21^\text{st}$ century. This approach largely limits learners' learning efficiency, as it is unable to provide personalized and timely feedback to learners, and remains linear in organization regardless of the different strengths, weaknesses, goals, and interests of different learners.
Recent developments in \emph{machine learning}-based personalized learning systems (PLSs) provide great potential to achieve personalized learning, by automatically mining data from learner interactions with educational content in order to provide a scalable, personalized education experience to a large number of learners (see \cite{itslesson,andes2005,knewton} for examples).

In our vision, a PLS should consist of two key components: 
\begin{itemize}
\item \emph{Learning analytics} (LA), which estimates the learners' knowledge states and dynamically traces their change over time, as the learners either \emph{learn} by interacting with learning materials (including \emph{learning resources}, e.g., textbook sections, lecture videos, and assessments, namely \emph{questions} in weekly quizzes, homework assignments, and exams), or \emph{forget} by not doing remedial studies (see, e.g., \cite{forget}).
\item \emph{Content analytics} (CA), which provides insight on the quality and content organization of all learning resources and the nature of the forgetting effect, as well as the difficulty and content organization of the available questions.
\end{itemize}

\subsection{SPARFA: sparse factor analysis for learning and content analytics}
\label{sec:sparfa}
The recently proposed sparse factor analysis (SPARFA) framework for personalized learning proposed statistical models and algorithms for machine learning-based LA and CA (\cite{sparfa}). SPARFA assumes that the learners' responses to questions in the domain of a course/assessment are governed by their knowledge on a small number of latent ``concepts.''
In particular, SPARFA relies on the following probability model for learners' graded responses to questions:
\begin{align*} 
Y_{i,j} & \sim \textit{Ber}(\Phi(Z_{i,j})) \quad \text{and} \quad Z_{i,j} = \vecw_i^T \vecc_j - \mu_i.
\end{align*}
Here, $Y_{i,j}$ is the binary-valued graded response of learner $j$ to question $i$, which is assumed to be a Bernoulli random variable, and $Z_{i,j}$ is a slack variable governing the probability of learner $j$ answering question $i$ correctly or incorrectly. $\Phi(\cdot)$ represents the inverse logit/probit link function. The variable $Z_{i,j}$ depends on three factors: (i) the question--concept association vector $\vecw_i$ which characterizes how question $i$ relates to each concept, (ii) the learner concept knowledge vector $\vecc_j$ of learner $j$, and (iii) the intrinsic difficulty parameter $\mu_i$ of question $i$.

The SPARFA framework jointly estimates (i) the question--concept association of each question, (ii) the concept knowledge of each learner, and (iii) the intrinsic difficulty of each question, solely from observed binary-valued (correct/incorrect) graded learner responses to questions, under assumptions (A1)--(A3).

This paper makes two major extensions to the SPARFA framework.
First, the SPARFA framework assumes that the learners' concept knowledge states remain \emph{constant} throughout the course/assessment. This assumption prohibits SPARFA from situations where learners' responses are made at different time instances, as is usual for homework sets assigned during a semester-long course. Second, the SPARFA framework only analyzes \emph{questions}, which measure learner knowledge states, but does not analyze learner \emph{knowledge state transitions}, which are induced by interacting with {learning resources} or by forgetting (\cite{retention1,retention2}). The analysis of learning resources is of utmost importance for a PLS, since this information enables the system to automatically recommend new resources to individual learners for remedial studies.

\subsection{SPARFA-Trace: time-varying learning and content analytics}
\label{sec:contributions}

\begin{figure}[t]
\centering
\includegraphics[width=1\columnwidth]{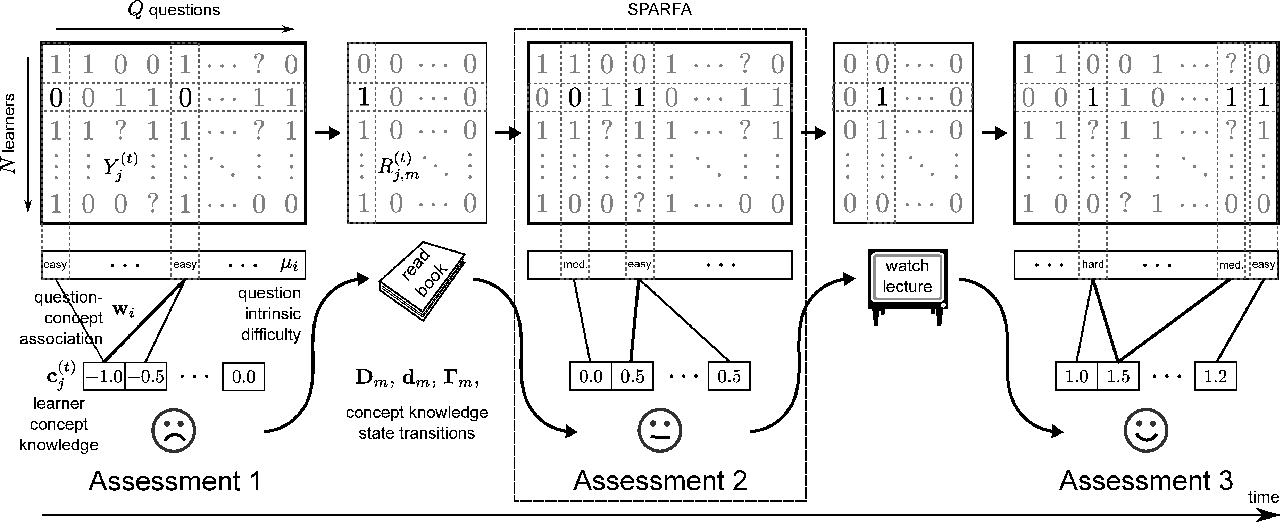} 
\vspace{-0.4cm}
\caption{The SPARFA-Trace framework processes the binary-valued graded learner response matrix $\bY$ (binary-valued, with $1$ denoting a correct response and $0$ denoting an incorrect response) and the learner activity matrices $\{\bR^{(t)}\}$ (binary-valued, with $1$ denoting a learner studied a learning resource) to (i) trace learner concept knowledge states $\vecc_j^{(t)}$ over time and (ii)  estimate learning resource content organization and quality parameters $\bD_m$, $\vecd_m$ and $\boldsymbol{\Gamma}_m$, together with question--concept association and question intrinsic difficulty parameters $\vecw_i$ and $\mu_i$, respectively.} 
\label{fig:graphextv}
\end{figure}

In this paper, we propose \emph{SPARFA-Trace}, a blind approximate Kalman filtering approach to perform joint \emph{time-varying} LA and CA. The main working principle of the approach is illustrated in \fref{fig:graphextv}.
Time-varying LA is performed by tracing each learner's concept knowledge (i.e., tracking the evolution of each learner's concept knowledge state vector $\vecc_j^{(t)}$ over different time instances $t$), based on (i) observed binary-valued (correct/incorrect) graded learner responses to questions matrix $\bY$ and (ii) available learner activity matrices $\bR^{(t)}$, as shown in \fref{fig:graphextv}.
CA is performed by estimating all learner concept knowledge state transition parameters $\bD_m$, $\vecd_m$ and $\boldsymbol{\Gamma}_m$, and question--concept associations and question intrinsic difficulties $\vecw_i$ and $\mu_i$, based on the estimated learner concept knowledge states at all time instances, as shown in \fref{fig:graphextv}.

In order to perform all these tasks, we extend the SPARFA framework developed in \cite{sparfa} using a new statistical model for learner knowledge state transitions that are induced by studying learning resources or the forgetting effect.
Armed with this model, LA corresponds to the task of estimating latent concept knowledge states in a dynamical system, where  binary-valued graded learner responses to questions are its observations.
We develop a message passing-based \emph{approximate} Kalman filtering approach to estimate the latent learner concept knowledge states at every time instance, as the underlying dynamical system is non-linear and non-Gaussian due to the binary-valued right/wrong observations, which inhibits the use of traditional Kalman filtering methods.
We also propose novel convex optimization-based algorithms within the expectation-maximization (EM) framework for CA, i.e., algorithms that estimate learners' concept knowledge state transition parameters and question-dependent parameters. The estimation of these parameters is crucial, as the approximate Kalman filtering approach requires all these parameters to be \emph{given}, which is, in general, not the case in real educational scenarios. 

To test and validate its effectiveness, we evaluate SPARFA-Trace on synthetic datasets and real-world educational datasets collected via OpenStax Tutor (\cite{ost}). 
SPARFA-Trace is accurate in tracing learner concept knowledge, and estimating learner concept knowledge state transition parameters and question-dependent parameters. Further, it outperforms existing approaches on predicting unobserved learner responses. 
We also demonstrate that SPARFA-Trace enables a PLS to (i) trace the learner concept knowledge state evolution over time in order to provide timely feedback to learners, and (ii) analyze the quality and content organization of all learning resources and assessment questions, in order to make effective and computerized recommendations to learners for remedial studies.

\subsection{Related work}
\label{sec:introrw}

The closest related work to SPARFA-Trace is knowledge tracing (KT).
KT is a popular technique for tracing learner knowledge evolution over time and for predicting future learner performance (see, e.g., \cite{kt,kt1,ktpardos}). Powerful as it is, KT suffers from the following drawbacks: 
(i) KT uses binary learner knowledge state representations, characterizing learners as whether they have mastered a certain concept (or skill) or not. The limited explanatory power of binary concept knowledge state representations prohibit the design of more powerful and sophisticated LA and CA algorithms.
(ii) KT assumes that each question is associated with exactly one concept. This restriction limits KT to very narrow educational domains (e.g., basic algebra), preventing it from generalizing to courses/assessments involving multiple concepts. 
(iii) KT uses a single ``probability of learning'' parameter to characterize the learner knowledge state transitions over time, and assumes that a concept cannot be forgotten once it is mastered. This modeling limitation forces all learner knowledge state transitions to be characterized by the same ``probability of learning'' parameter, prohibiting KT from performing CA, i.e., analyzing the quality and content organization of different learning resources that lead to different learner knowledge state transitions.
See \fref{sec:rlwork} for detailed discussions and comparisons with previous work in KT and other machine learning-based approaches to  personalized learning.

\subsection{Paper organization}

In \fref{sec:model}, we introduce the SPARFA-Trace statistical model for learner knowledge state transitions that are induced by interacting with learning resources or forgetting. 
In \fref{sec:kf}, we detail the approximate Kalman filtering approach for learner concept knowledge tracing.
In \fref{sec:parameter}, we detail convex optimization-based algorithms to estimate learners' concept knowledge state transition parameters and question-dependent parameters. 
In \fref{sec:sims}, we evaluate SPARFA-Trace on synthetic and real-world educational datasets. 
In \fref{sec:rlwork}, we provide a brief overview of related KT and machine learning-based techniques for personalized learning. We conclude in \fref{sec:conclusions}.

\section{Statistical Model for Time-Varying Learning and Content Analytics}
\label{sec:model}

We start by extending the SPARFA framework (\cite{sparfa}) to trace learner concept knowledge over time, and propose the corresponding statistical model in \fref{sec:modelqa}.
In \fref{sec:modeltrans}, we then characterize the transition of a learner's concept knowledge states between consecutive time instances as an affine model, which is parameterized by (i) the particular learning resource(s) the learner interacted with, and (ii) how these learning resource(s) affect learners' concept knowledge states.

\subsection{Statistical model for time-varying graded learner responses to questions}
\label{sec:modelqa}

The proposed statistical model characterizes the probability that a learner answers a question correctly at a particular time instance in terms of: (i) the learner's knowledge on every concept at this particular time instance, (ii) how the question relates to each concept, and (iii) the intrinsic difficulty of the question. 
To this end, let $N$ denote the number of learners, $K$ the number of latent concepts in the course/assessment, and $T$ the total number of time instances throughout the course/assessment. 
We define the $K$-dimensional vectors $\vecc_j^{(t)} \in \mathbb{R}^K, t \in \{1, \ldots, T\}, j \in \{1, \ldots, N \}$, to represent the latent concept knowledge state of the $j^\text{th}$ learner at time instance $t$.
Let $Q$ be the total number of questions. We further define the mapping $i(t,j): \{1, \ldots, T\} \times \{1, \ldots, N \}  \mapsto \{1, \ldots, Q\}$, which maps learner and time instance indices to question indices; this information can  be extracted from the learner activity log. We will use the shorthand notation $i^{(t)}_j = i(t,j)$ to denote the index of the question that the $j^\text{th}$ learner answers $i^{(t)}_j$ at time instance $t$. 
Under this notation, we define the $K$-dimensional vector $\vecw_{i^{(t)}_j} \in \mathbb{R}^K, i \in \{1, \ldots, Q\}$ as the question--concept association vector of the question that the $j^\text{th}$ learner answered at time instance $t$. Finally, we define the scalar $\mu_{i^{(t)}_j} \in \mathbb{R}$ to be the intrinsic difficulty of question $i_j^{(t)}$, with a large, positive values of $\mu_{i^{(t)}_j}$ representing a difficult question, while a small, negative values of $\mu_{i^{(t)}_j}$ represent an easy one.

Given these quantities, we characterize the binary-valued graded response, where $1$ denotes a correct response and $0$ an incorrect response, of learner $j$ to question $i^{(t)}_j$ at time instance $t$ as a Bernoulli random variable:
\begin{align} 
Y_j^{(t)} & \sim \textit{Ber}(\Phi(Z_j^{(t)})), \,\,\,\quad (t,j) \in \Omega_\text{obs}, \notag \\
Z_j^{(t)} & = \vecw_{i^{(t)}_j}^T \vecc_j^{(t)} - \mu_{i^{(t)}_j},  \quad \forall t,j. \label{eq:qa}
\end{align}
Here, the set $\Omega_\text{obs}\subseteq\{1,\ldots,T\}\times\{1,\ldots,N\}$ contains the indices associated with the observed graded learner response data, since some learner responses might not be observed in practice. 
$\Phi(z)$ denotes the \emph{inverse probit link} function
$\Phi_\text{pro}(z) = \int_{-\infty}^z \! \mathcal{N}(t) \, \mathrm{d}t$,
where $\mathcal{N}(t) = \frac{1}{\sqrt{2 \pi}} \exp(-t^2/2)$ is the probability density function (PDF) of the standard normal distribution. (Note that the inverse logit link function could also be used. However, the inverse probit link function simplifies the calculations detailed in \fref{sec:akf}.) The likelihood of an observation $Y_j^{(t)}$ can, alternatively, be written as 
\begin{align*}
p(Y_j^{(t)} \!\mid\! \vecc_j^{(t)}) = \Phi\big((2Y_j^{(t)} -1)(\vecw_{i^{(t)}_j}^T \vecc_j^{(t)} - \mu_{i^{(t)}_j})\big),
\end{align*}
an expression that we will often use in the remainder of the paper.

Following the original SPARFA framework, we impose the following model assumptions:
\begin{itemize}
\item[(A1)] \emph{The number of concepts is much smaller than the number of questions and the number of learners}: This assumption imposes a low-dimensional model on the learners' responses to questions.
\item[(A2)] \emph{The vector $\vecw_i$ is sparse}: This assumption bases on the observation that each question should only be associated with a few concepts out of all concepts in the domain of a course/assessment.
\item[(A3)] \emph{The vector $\vecw_i$ has non-negative entries}: This assumption enables one to interpret the entries in $\vecc_j$ to be the latent concept knowledge of each learner, with positive values represent high concept knowledge, while negative values represent low concept knowledge.
\end{itemize}

These assumptions are reasonable in the majority of real-world educational scenarios, alleviating the common identifiability issue (i.e., if $Z_{i,j} =  \vecw_i^T \vecc_j$, then for any orthonormal matrix $\bQ$ with $\bQ^T \bQ = \bI$, we have $Z_{i,j} = \vecw_i^T \bQ^T \bQ \vecc_j=\widetilde{\vecw}_i^T \widetilde{\vecc}_j$. Hence, estimating $\vecw_i$ and $\vecc_j$ is, in general, non-unique up to a unitary unitary transformation. See \cite{factoran} and \cite{sparfa} for more details) of factor analysis and improving the interpretability of the variables $\vecw_i$, $\vecc_j$, and~$\mu_i$.

\subsection{Statistical model for learner knowledge state transition}
\label{sec:modeltrans}

The SPARFA model \fref{eq:qa} assumes that each learner's concept knowledge remains \emph{constant} throughout a course/assessment. 
Although this assumption is valid in the setting of a single test or exam, it provides limited explanatory power in analyzing the (possibly semester-long) process of a course, during which the learners' concept knowledge evolves through time. 
Concept knowledge state evolution can happen due to the following reasons: (i) A learner can interact with learning resources (read a section of an assigned textbook, watch a lecture video, conduct a lab experiment, or do a computer simulation), all likely to result in an increase of their concept knowledge. (ii) A learner can simply forget a learned concept, resulting in a decrease of their concept knowledge. For the sake of simplicity of exposition, we will treat the forgetting effect (\cite{forget}) as a special learning resource that reduces learners' concept knowledge over time. 

We  propose a latent state transition model that models learner concept knowledge evolution between two consecutive time instances. To this end, we assume that there is a total number of $M$ distinct learning resources. We define the mapping $m(t,j): \{1, \ldots, T\} \times \{1, \ldots, N \} \mapsto \{1, \ldots, M\}$ from time and learner indices to learning resource indices; this information can be extracted from the learner activity log. We will use the shorthand notation $m_j^{(t-1)} = m(t-1,j)$ to denote the index of the learning resource that learner $j$ studies between time instance $t-1$ and time instance $t$. Under these notations, the learner activity summary matrices $\bR^{(t)}$ illustrated in \fref{fig:graphextv} are defined by $R_{j,m_j^{(t)}}^{(t)}  = 1,\; \forall (t, j)$, meaning that learner $j$ interacted with learning resource $m_j^{(t)}$ at time instance $t$, and $0$ otherwise.

We are now ready to model the transition of learner $j$'s latent concept knowledge state from time instance $t-1$ to $t$ as:
\begin{align}
\vecc_j^{(t)} = (\bI_K + \bD_{m_j^{(t-1)}}) \vecc_j^{(t-1)} + \vecd_{m_j^{(t-1)}} + \boldsymbol{\epsilon}_j^{(t-1)}, \quad \boldsymbol{\epsilon}_j^{(t-1)} \distas \mathcal{N}(\boldsymbol{0}_K, \boldsymbol{\Gamma}_{m_j^{(t-1)}}), \label{eq:trans}
\end{align}
where $\bI_K$ is the $K \times K$ identity matrix;
$\bD_{m_j^{(t-1)}}$, $\vecd_{m_j^{(t-1)}}$ and $\boldsymbol{\Gamma}_{m_j^{(t-1)}}$ are latent learner concept knowledge state transition parameters, which define an affine model on the transition of the $j^\text{th}$ learner's concept knowledge state by interacting with learning resource $m_j^{(t-1)}$ between time instances $t-1$ and $t$. $\bD_{m_j^{(t-1)}}$ is a $K \times K$ matrix, $\vecd_{m_j^{(t-1)}}$ is a $K \times 1$ vector, and $\boldsymbol{0}_K$ is the $K$-dimensional zero vector.
The covariance matrix $\boldsymbol{\Gamma}_{m_j^{(t-1)}}$ characterizes the uncertainty induced in the learner concept knowledge state transition by interacting with learning resource $m_j^{(t-1)}$. 
Note that~\fref{eq:trans} also has the following equivalent form:
\begin{align}
p(\vecc_j^{(t)} \!\mid\! \vecc_j^{(t-1)}) = \mathcal{N} \! \left( \vecc_j^{(t)} \!\mid\! (\bI_K +\bD_{m_j^{(t-1)}}) \vecc_j^{(t-1)} + \vecd_{m_j^{(t-1)}}, \; \boldsymbol{\Gamma}_{m_j^{(t-1)}} \right), \label{eq:transm}
\end{align}
where $\mathcal{N}(\vecx | \boldsymbol{\mu}, \boldsymbol{\Sigma})$ represents a multivariate Gaussian distribution with mean vector $\boldsymbol{\mu}$ and covariance matrix $\boldsymbol{\Sigma}$.

In order to reduce the number of parameters, to improve identifiability of the parameters $\bD_{m_j^{(t-1)}}$, $\vecd_{m_j^{(t-1)}}$ and $\boldsymbol{\Gamma}_{m_j^{(t-1)}}$, and to account for real-world educational scenarios, we impose three additional assumptions on the learner knowledge state transition matrix $\bD_{m_j^{(t-1)}}$:
\begin{itemize}
\item[(A4)] \emph{$\bD_{m_j^{(t-1)}}$ is lower triangular}: 
This assumption means that, the $k^\text{th}$ entry in the learner concept knowledge vector $\vecc_j^{(t)}$ is only influenced by the the $1^\text{st}, \ldots, (k-1)^\text{th}$ entry in $\vecc_j^{(t-1)}$. 
As a result, the upper entries in $\vecc_j^{(t-1)}$ represent pre-requisite concepts that are covered early in the course, while lower entries represent advanced concepts that are covered towards the end of the course. Using this assumption, it is possible to extract prerequisite relationships among concepts purely from learner response data.

\item[(A5)] \emph{$\bD_{m_j^{(t-1)}}$ has non-negative entries}: This assumption ensures that having low concept knowledge at time instance $t-1$ (negative entries in $\vecc_j^{(t-1)}$) does not result in high concept knowledge at time instance $t$ (positive entries in $\vecc_j^{(t)}$). 

\item[(A6)] \emph{$\bD_{m_j^{(t-1)}}$ is sparse}: This assumption amounts for the observation that learning resources typically only cover a small subset of concepts among all concepts covered in a course.

\end{itemize}

In contrast to the learner concept knowledge transition matrix $\bD_{m_j^{(t-1)}}$, we do not impose sparsity or non-negativity properties on the intrinsic learner concept knowledge state transition vector $\vecd_{m_j^{(t-1)}}$ in \fref{eq:trans}; 
large, positive values in $\vecd_{m_j^{(t-1)}}$ represent learning resources with good quality that boost learners' concept knowledge, while small, negative values in $\vecd_{m_j^{(t-1)}}$ represent learning resources that reduce learners' concept knowledge.
This setting enables our framework to model cases of poorly designed, misleading, or off-topic learning resources that distract or confuse learners. Note that the forgetting effect can also be modeled as a learning resource with negative entries in $\vecd_{m_j^{(t-1)}}$. 

To further reduce the number of parameters, the covariance matrix $\boldsymbol{\Gamma}_{m_j^{(t-1)}}$ is assumed to be diagonal, implying that the uncertainties of learning resources on learners' knowledge states are not correlated among different concepts. This assumption is mainly made for simplicity; the analysis of more evolved models is left for future work.

In the next section, we will describe how to estimate the learners' concept knowledge state vectors $\vecc_j^{(t)}, \, \forall t,j$, given observed graded learner responses $\bY_j^{(t)}, (t,j) \in \Omega_\text{obs}$, and all parameters $\bD_{m_j^{(t-1)}}, \vecd_{m_j^{(t-1)}}, \boldsymbol{\Gamma}_{m_j^{(t-1)}}, \vecw_{i_j^{(t)}}, \mu_{i_j^{(t)}}, \, \forall {t,j}$, i.e., how to perform time-varying learning analytics. 
Then, in \fref{sec:parameter}, we will introduce methods to estimate these parameters and thus analyze the quality and content organization of all learning resources and questions, i.e., performing \emph{content analytics}.

\section{Time-Varying Learning Analytics}
\label{sec:kf}

We now introduce a message passing-based approximate Kalman filtering approach for learner concept knowledge tracing. 
Since the observed data is binary-valued graded learner responses to questions, we cannot simply use common Kalman filter methods that assume Gaussian observation models. 
We start with a brief review of Kalman filtering and smoothing, and then introduce the necessary approximations in the Kalman filtering approach to estimate latent learner concept knowledge states at all time instances.
For simplicity, we will drop the learner index $j$ in this section, i.e., quantities $\bD_{m_j^{(t-1)}}$ and $\vecd_{m_j^{(t-1)}}$ are replaced by $\bD_{m^{(t-1)}}$ and $\vecd_{m^{(t-1)}}$.
Moreover, we use the shorthand notation $\overline{\bD}_{m^{(t-1)}}$ for the quantity $\bI_{K} + \bD_{m^{(t-1)}}$.

\subsection{Kalman filtering}
\label{sec:oldkf}

Kalman filtering (\cite{kalman,kfbook}) solves a key inference problem in linear dynamical systems (LDS), where the system consist of a series of continuous latent state variables with Markovian state transition property and a Gaussian observation model. The following derivations briefly summarize the results in~\cite{minka}. 

\begin{figure}
\vspace{-0cm}
\center
\includegraphics[width=.84\columnwidth]{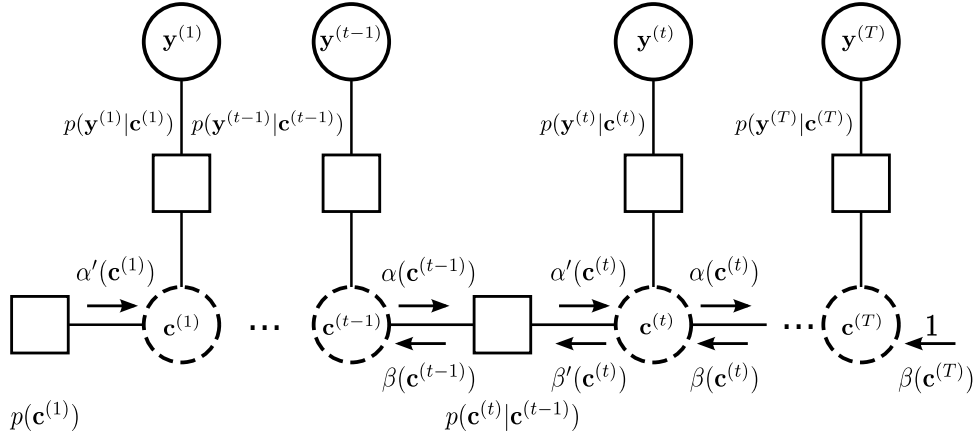}
\vspace{-0.0cm}
\caption{Factor graph message passing scheme for the inference of a set of $T$ latent state variables with Markovian transition properties from (possibly noisy) observations. \label{fig:lds}}
\end{figure}

The Markov chain consist of a series of $T$ latent state variables $\vecc^{(t)}, t = 1,\ldots,T$, and observations $\vecy^{(t)}, t = 1,\ldots,T$. 
Due to the Markovian property of the system, the joint probability of all latent states and all observations can be factorized as
\begin{align*}
p(\vecc^{(1)}, \ldots, \vecc^{(T)}, \vecy^{(1)}, \ldots, \vecy^{(T)}) = p(\vecc^{(1)}) \, p(\vecy^{(1)} \!\mid\! \vecc^{(1)}) \textstyle \prod_{t=2}^{T} p(\vecc^{(t)} \!\mid\! \vecc^{(t-1)}) \, p(\vecy^{(t)} \!\mid\! \vecc^{(t)}).
\end{align*}
A visualization of the dynamical system as a factor graph (\cite{factorgraph1,factorgraph}) is shown in \fref{fig:lds}.

The inference algorithm, which estimates the vectors $\vecc^{(t)}, \, \forall t$, based on the observations $\vecy^{(t)}, \, \forall t$, consist of two parts. First, a forward message passing phase (i.e., the Kalman filtering phase) is performed. Then, using estimates obtained during the Kalman filtering phase, a backward message passing phase (often referred to as Kalman smoothing or Rauch-Tung-Streibel (RTS) smoothing) is also performed.

In the forward message passing phase, the goal is to estimate latent state variables $\vecc^{(t)}$ based on the previous observations $\vecy^{(1)}, \ldots, \vecy^{(t)}$. In other words, the value of interest is $p(\vecc^{(t)} \!\mid\! \vecy^{(1)}, \ldots, \vecy^{(t)}), \, \forall t$. 
This quantity can be obtained via a message passing scheme outlined in \fref{fig:lds}. Specifically, by starting at $t=1$, the incoming message to variable node $\vecc^{(1)}$ is given by
$\alpha'(\vecc^{(1)}) = p(\vecc^{(1)})$.
The outgoing message from variable node $\vecc^{(1)}$ to factor node $p(\vecc^{(2)} \!\mid\! \vecc^{(1)})$ is then given by
\begin{align*}
\alpha(\vecc^{(1)}) = \alpha'(\vecc^{(1)}) \, p(\vecy^{(1)} \!\mid\! \vecc^{(1)})= p(\vecc^{(1)}) \, p(\vecy^{(1)} \!\mid\! \vecc^{(1)}) = b_1 \, p(\vecc^{(1)} \!\mid\! \vecy^{(1)}),
\end{align*}
according to Bayes rule, where $b_1 = p(\vecy^{(1)})$ is the \emph{scaling factor}. 
Recursively following these rules, the outgoing message $\alpha(\vecc^{(t-1)})$ from variable node $\vecc^{(t-1)}$ to the factor node $p(\vecc^{(t)} \!\mid\! \vecc^{(t-1)})$ at time $t$ is given by
\begin{align*}
\alpha(\vecc^{(t-1)}) = \left( \textstyle \prod_{\tau = 1}^{t-1} b^{(\tau)} \right) p(\vecc^{(t-1)} \!\mid\! \vecy^{(1)}, \ldots, \vecy^{(t-1)}).
\end{align*}
The outgoing message $\alpha'(\vecc^{(t)})$ from factor node $p(\vecc^{(t)} \!\mid\! \vecc^{(t-1)})$ to variable node $\vecc^{(t)}$ is given by
\begin{align*}
\alpha'(\vecc^{(t)}) & = \int \alpha(\vecc^{(t-1)}) p(\vecc^{(t)} \!\mid\! \vecc^{(t-1)}) \mathrm{d}\vecc^{(t-1)}
 = \left( \textstyle \prod_{\tau = 1}^{t-1} b^{(\tau)} \right) \, p(\vecc^{(t)} \!\mid\! \vecy^{(1)}, \ldots, \vecy^{(t-1)}).
\end{align*}
The outgoing message $\alpha(\vecc^{(t)})$ from variable node $\vecc^{(t)}$ is given by
\begin{align*}
\alpha(\vecc^{(t)}) & = \alpha'(\vecc^{(t)}) \, p(\vecy^{(t)} \!\mid\! \vecc^{(t)}) 
 = \left( \textstyle \prod_{\tau = 1}^{t} b^{(\tau)} \right) p(\vecc^{(t)} \!\mid\! \vecy^{(1)}, \ldots, \vecy^{(t)}), 
\end{align*}
where $b^{(t)} = p(\vecy^{(t)} \!\mid\! \vecy^{(1)}, \ldots, \vecy^{(t-1)})$. 
We can see that a scaled version of $\alpha(\vecc^{(t)})$, $\widehat{\alpha}(\vecc^{(t)}) = \frac{\alpha(\vecc^{(t)})}{\textstyle \prod_{\tau = 1}^{t} b^{(\tau)}} = p(\vecc^{(t)} \!\mid\! \vecy^{(1)}, \ldots, \vecy^{(t)})$, is exactly the value of interest.

The derivations above show that $\widehat{\alpha}(\vecc^{(t)})$ can be obtained in recursive fashion via
\begin{align}
\label{eq:frecursion}
b^{(t)} \, \widehat{\alpha}(\vecc^{(t)}) = p(\vecy^{(t)} \!\mid\! \vecc^{(t)}) \int p(\vecc^{(t)} \!\mid\! \vecc^{(t-1)}) \,\widehat{\alpha}(\vecc^{(t-1)}) \mathrm{d}\vecc^{(t-1)}.
\end{align}
The key for obtaining a tractable and efficient estimator for $p(\vecc^{(t)} \!\mid\! \vecy^{(1)}, \ldots, \vecy^{(t)})$ is that the transition probability $p(\vecc^{(t)} \!\mid\! \vecc^{(t-1)})$ and the observation likelihood $p(\vecy^{(t)} \!\mid\! \vecc^{(t)})$ satisfy certain properties such that the messages $\widehat{\alpha}(\vecc^{(t)})$ and $\widehat{\alpha}(\vecc^{(t-1)})$ take on the same functional form, just with different parameters. 
A LDS is a special case in which the transition probability and the observation likelihood are (multivariate) Gaussians of are of the following form:
\begin{align*}
& p(\vecc^{(t)} \!\mid\! \vecc^{(t-1)}) = \mathcal{N} (\vecc^{(t)} \!\mid\! \overline{\bD}_{m^{(t-1)}} \vecc^{(t-1)} + \vecd_{m^{(t-1)}}, \boldsymbol{\Gamma}_{m^{(t-1)}}), \\
& p(\vecy^{(t)} \!\mid\! \vecc^{(t)}) = \mathcal{N} (\vecy^{(t)} \!\mid\! \bW_{i^{(t)}} \vecc^{(t)}, \boldsymbol{\Sigma}_{i^{(t)}}).
\end{align*}
Here, $\boldsymbol{\Gamma}_{m^{(t-1)}}$ is the covariance matrix for state transition, $\bW_{i^{(t)}}$ is the measurement matrix, and $\boldsymbol{\Sigma}_{i^{(t)}}$ is the covariance matrix for the multivariate observation of the system.
In order for the functional form of the messages to stay the same over time, the messages are also Gaussian, i.e., $\widehat{\alpha}(\vecc^{(t)}) \distas \mathcal{N} (\vecc^{(t)} \!\mid\! \vecm^{(t)}, \bV^{(t)})$.
Under these conditions, the forward message passing recursion \fref{eq:frecursion} takes on a compact form
\begin{align}
\label{eq:fp}
b^{(t)} \, \widehat{\alpha}(\vecc^{(t)}) = \mathcal{N}\! \left(\vecc^{(t)} \!\mid\! \vecm^{(t)}, \bV^{(t)}\right),
\end{align}
with the parameters $b^{(t)}$, $\vecm^{(t)}$ and $\bV^{(t)}$ given by
\begin{align*}
 \vecm^{(t)} & = \overline{\bD}_{m^{(t-1)}} \vecm^{(t-1)} + \vecd_{m^{(t-1)}} + \bK^{(t)} \left(\vecy^{(t)} - \bW_{i^{(t)}} \left(\overline{\bD}_{m^{(t-1)}} \vecm^{(t-1)} + \vecd_{m^{(t-1)}}\right)\right), \\
 \bV^{(t)} & = \left( \bI - \bK^{(t)} \bW_{i^{(t)}} \right) \bP^{(t-1)}, \\
 b^{(t)} & = \mathcal{N} \left( \vecy^{(t)} \!\mid\! \bW_{i^{(t)}} \left( \overline{\bD}_{m^{(t-1)}} \vecm^{(t-1)} + \vecd_{m^{(t-1)}}\right), \bW_{i^{(t)}} \bP^{(t-1)} \bW_{i^{(t)}}^T + \boldsymbol{\Sigma}_{i^{(t)}}  \right),
\end{align*}
in which the matrices $\bK^{(t)}$ and $\bP^{(t-1)}$ are given by
\begin{align*}
\bK^{(t)} & = \bP^{(t-1)} \bW_{i^{(t)}}^T \left(\bW_{i^{(t)}} \bP^{(t-1)} \bW_{i^{(t)}}^T + \boldsymbol{\Sigma}_{i^{(t)}} \right)^{-1}, \\
\bP^{(t-1)} & = \overline{\bD}_{m^{(t-1)}} \bV^{(t-1)} \overline{\bD}_{m^{(t-1)}}^T + \boldsymbol{\Gamma}_{m^{(t-1)}}.
\end{align*}
The recursion starts with a prior $p(\vecc^{(1)})   = \mathcal{N} (\vecc^{(1)} \!\mid\! \vecm^{(0)}, \bV^{(0)})$, and 
\begin{align*}
\vecm^{(1)}& = \vecm^{(0)} + \bK^{(1)} \left( \vecy^{(1)} - \bW_{i^{(1)}} \vecm^{(0)}\right), \\
\bV^{(1)} & = \left( \bI_{K} - \bK^{(1)} \bW_{i^{(1)}} \right) \bV^{(0)}, \\
\bK^{(1)} & = \bV^{(0)} \bW_{i^{(1)}}^T \left( \bW_{i^{(1)}} \bV^{(0)} \bW_{i^{(1)}}^T + \boldsymbol{\Sigma}_{i^{(1)}} \right)^{-1}, \\
b^{(1)} & = \mathcal{N} \left( \vecy^{(1)} \!\mid\! \bW_{i^{(1)}} \vecm^{(0)}, \bW_{i^{(1)}} \bV^{(0)} \bW_{i^{(1)}}^T + \boldsymbol{\Sigma}_{i^{(1)}} \right).
\end{align*}
The initial prior mean and variance for $\vecc^{(1)}$ are assumed to be $\vecm^{(0)} = \boldsymbol{0}_{K}$ and $\bV^{(0)} = \sigma_0^2 \bI_K$ in what follows.

\subsection{Kalman smoothing}
\label{sec:oldks}

As detailed above, Kalman filtering can be utilized to obtain $p(\vecc^{(t)} \!\mid\! \vecy^{(1)}, \ldots, \vecy^{(t)})$, an estimate on the latent state at time instance $t$, given all observations $\vecy^{(\tau)}$ for $\tau < t$. 
This estimate is the value of interest for a variety of real-time tracking applications, since decisions have to be made based on all available observations up to a certain time instance. However, in our application, one could also use  observations at $\tau \geq t$ to obtain a better estimate of the latent state at time instance $t$.
In other words, the value of interest is now $p(\vecc^{(t)} \!\mid\! \vecy^{(1)}, \ldots, \vecy^{(T)})$. 
In order to estimate this value, a set of backward recursions similar to the set of forward recursions \fref{eq:frecursion} can be used.

Following the backward message passing scheme described in \cite{factorgraph}, the backwards message starts with a ''one'' message going into variable node $\vecc^{(T)}$: $\beta (\vecc^{(T)}) = \boldsymbol{1}$ (as shown in \fref{fig:lds}).
Then, the outgoing message from variable node $\vecc^{(T)}$ into factor node $p(\vecc^{(T)} \!\mid\! \vecc^{(T-1)})$ is
\begin{align*}
\beta'(\vecc^{(T)}) = p(\vecy^{(T)} \!\mid\! \vecc^{(T)}),
\end{align*}
and the outgoing message from factor node $p(\vecc^{(T)} \!\mid\! \vecc^{(T-1)})$ into variable node $\vecc^{(T-1)}$ is
\begin{align*}
\beta(\vecc^{(T-1)}) & = \int p(\vecc^{(T)} \!\mid\! \vecc^{(T-1)}) p(\vecy^{(T)} \!\mid\! \vecc^{(T)}) \mathrm{d}\vecc^{(T)}
 = p(\vecy^{(T)} \!\mid\! \vecc^{(T-1)}).
\end{align*}
Following this convention, we obtain the following recursion:
\begin{align*}
\beta(\vecc^{(t-1)}) & = \int p(\vecc^{(t)} \!\mid\! \vecc^{(t-1)}) p(\vecy^{(t)} \!\mid\! \vecc^{(t)}) \beta(\vecc^{(t)}) \mathrm{d}\vecc^{(t)}
 = p(\vecy^{(t)}, \ldots, \vecy^{(T)} \!\mid\! \vecc^{(t-1)}),
\end{align*}
where we have implicitly used the Markovian properties of the latent state variables. 
Now, the marginal distribution of latent state variables $\vecc^{(t)}$ can be written as a product of the incoming messages into variable node $\vecc^{(t)}$ from both forward and backward recursions, i.e.,
\begin{align*}
p(\vecc^{(t)} \!\mid\! \vecy^{(1)}, \ldots, \vecy^{(T)}) & = \frac{p(\vecc^{(t)} \!\mid\! \vecy^{(1)}, \ldots, \vecy^{(T)}) \, p(\vecy^{(t+1)}, \ldots, \vecy^{(T)} \!\mid\! \vecy^{(1)}, \ldots, \vecy^{(t)})}{p(\vecy^{(t+1)}, \ldots, \vecy^{(T)} \!\mid\! \vecy^{(1)}, \ldots, \vecy^{(t)})} \!
 = \widehat{\alpha}(\vecc^{(t)}) \widehat{\beta}(\vecc^{(t)}),
\end{align*}
where $\widehat{\beta}(\vecc^{(t)}) = \frac{\beta(\vecc^{(t)})}{\textstyle \prod_{\tau = t+1}^{T} b^{(\tau)}}$ is a scaled version of $\beta(\vecc^{(t)})$. 
Now, the backward recursion is as follows:
\begin{align}
\label{eq:brecursion}
b^{(t)} \, \widehat{\beta}(\vecc^{(t-1)}) & = \int_{\vecc^{(t)}} p(\vecc^{(t)} \!\mid\! \vecc^{(t-1)}) p(\vecy^{(t)} \!\mid\! \vecc^{(t)}) \widehat{\beta}(\vecc^{(t)}) \mathrm{d}\vecc^{(t)}.
\end{align}
Although it is possible to obtain a backward recursion for $\widehat{\beta}(\vecc^{(t)})$, the common approach uses a recursion directly on $\widehat{\alpha}(\vecc^{(t)})\widehat{\beta}(\vecc^{(t)})$ to obtain the value of interest $p(\vecc^{(t)} \!\mid\! \vecy^{(1)}, \ldots, \vecy^{(T)})$. By multiplying both sides of the equation \fref{eq:brecursion} by $\widehat{\alpha}(\vecc^{(t-1)})$, we obtain
\begin{align*}
\widehat{\alpha}(\vecc^{(t-1)})\widehat{\beta}(\vecc^{(t-1)}) = \widehat{\alpha}(\vecc^{(t-1)})  \int_{\vecc^{(t)}} p(\vecc^{(t)} \!\mid\! \vecc^{(t-1)}) p(\vecy^{(t)} \!\mid\! \vecc^{(t)}) \frac{\widehat{\alpha}(\vecc^{(t)})\widehat{\beta}(\vecc^{(t)})}{b^{(t)} \, \widehat{\alpha}(\vecc^{(t)})} \mathrm{d}\vecc^{(t)},
\end{align*}
which can be computed recursively as a backward message passing process, given the estimates \fref{eq:fp} following the completion of the forward message passing process detailed in \fref{sec:oldkf}. 
Specifically, for a LDS, the recursions take the form:
\begin{align}
\label{eq:bp}
\widehat{\alpha}(\vecc^{(t-1)})\widehat{\beta}(\vecc^{(t-1)}) = \mathcal{N} (\vecc^{(t-1)} \!\mid\! \widehat{\vecm}^{(t-1)}, \widehat{\bV}^{(t-1)})
\end{align}
with the parameters $\widehat{\vecm}^{(t-1)}$ and $\widehat{\bV}^{(t-1)}$ given by
\begin{align*}
\widehat{\vecm}^{(t-1)} & = \vecm^{(t-1)} + \bJ^{(t-1)} \left( \widehat{\vecm}^{(t)} - \overline{\bD}_{m^{(t-1)}} \vecm^{(t-1)} - \vecd_{m^{(t-1)}} \right), \\
\widehat{\bV}^{(t-1)} & = \bV^{(t-1)} + \bJ^{(t-1)}  \left( \widehat{\bV}^{(t)} - \bP^{(t-1)} \right)   (\bJ^{(t-1)})^T, \\
\bJ^{(t-1)} & = \bV^{(t-1)} (\overline{\bD}_{m^{(t-1)}})^T (\bP^{(t-1)})^{-1}.
\end{align*}
We start the recursion with $\widehat{\vecm}^{(T)} = \vecm^{(T)}$ and $\widehat{\bV}^{(T)} = \bV^{(T)}$, since $\beta( \vecc^{(T)}) = \boldsymbol{1}$.

In the above derivations, we have assumed that $\vecy^{(t)}$ is observed for all $t$. If $\vecy^{(t)}$ is unobserved, then the message passing scheme will simply have $\alpha(\vecc^{(t)}) = \alpha'(\vecc^{(t)})$ and $\beta'(\vecc^{(t)}) = \beta(\vecc^{(t)})$ instead, while the rest of the recursions remain unaffected.

\subsection{Approximate Kalman filtering for learner concept knowledge tracing}
\label{sec:akf}

The basic Kalman filtering and smoothing approaches (Equations~\fref{eq:fp} and \fref{eq:bp}) are only suitable for applications with a Gaussian latent state transition model and a Gaussian observation model, while the forward and backward recursions (Equations~\fref{eq:frecursion} and \fref{eq:brecursion}) holds for arbitrary state transition and observation models.
When attempting to trace latent learner concept knowledge states under the SPARFA model, it is not possible to make Gaussian observations of these states. Concretely, we have only binary-valued graded learner responses as our observations in the present application.
We will now detail approximations that have to be made to enable the estimation of latent learner concept knowledge states under our model.

As introduced in \fref{sec:model}, the observation model at time $t$ is given by \fref{eq:qa} and the state transition model is given by \fref{eq:transm}. 
Therefore, the recursion formula for the forward message passing process \fref{eq:frecursion} becomes
\begin{align}
\label{eq:arecursion}
\notag b^{(t)} \, \widehat{\alpha}(\vecc^{(t)}) & = p(Y^{(t)} \!\mid\! \vecc^{(t)}) \int p(\vecc^{(t)} \!\mid\! \vecc^{(t-1)}) \,\widehat{\alpha}(\vecc^{(t-1)}) \mathrm{d}\vecc^{(t)}\\
\notag & \! = \! \Phi \! \left( \! \left(2 Y^{(t)} -1 \right)\! \left(\! \vecw_{i^{(t)}}^T \vecc^{(t)}\! -\! \mu_{i^{(t)}}\! \right)\! \right) \!\int \!\mathcal{N}\left(\!\vecc^{(t)} \!\!\mid\! \!\overline{\bD}_{m^{(t-1)}} \vecc^{(t-1)} \!+\! \vecd_{m^{(t-1)}}, \!\boldsymbol{\Gamma}_{m^{(t-1)}} \!\right) \! \\
\notag & \quad \mathcal{N}\left(\!\vecc^{(t-1)}\! \!\mid\!\! \vecm^{(t-1)}, \!\bV^{(t-1)}\! \right) \mathrm{d}\vecc^{(t)}\\
\notag & \! = \! \Phi \! \left( \! \left(2 Y^{(t)} -1 \right)\! \left( \vecw_{i^{(t)}}^T \vecc^{(t)}\! -\! \mu_{i^{(t)}} \! \right) \! \right) \mathcal{N} \! ( \! \vecc^{(t)} \! \!\mid\! \! \overline{\bD}_{m^{(t-1)}} \vecm^{(t-1)} \!+\! \vecd_{m^{(t-1)}}, \\
\notag & \quad \overline{\bD}_{m^{(t-1)}} \bV^{(t-1)} \overline{\bD}_{m^{(t-1)}}^T \!+ \!\boldsymbol{\Gamma}_{m^{(t-1)}}) \\
& =  \! \Phi \! \left( \! \left(2 Y^{(t)} -1 \right)\! \left(\! \vecw_{i^{(t)}}^T \vecc^{(t)}\! -\! \mu_{i^{(t)}}\! \right)\! \right) \! \mathcal{N} \! \left( \! \vecc^{(t)} \! \!\mid\! \widetilde{\vecm}^{(t)}, \widetilde{\bV}^{(t)} \right), 
\end{align}
where we used a \emph{tilde} to denote the mean and covariance of the messages $\alpha'(\vecc^{(t-1)})$. 

Equation~\fref{eq:arecursion} shows that $\widehat{\alpha}(\vecc^{(t)})$ is no longer Gaussian even if $\widehat{\alpha}(\vecc^{(t-1)})$ is Gaussian, under the probit binary observation model.
Thus, the closed-form updates in~\fref{eq:fp} and \fref{eq:bp} can no longer be applied.
Therefore, we have to perform an approximate message passing approach within the Kalman filtering framework to arrive at a tractable estimator of $\vecc^{(t)}$. 
In order to do so, a number of approaches has been proposed to approximate $\widehat{\alpha}(\vecc^{(t)})$ by a Gaussian distribution $\mathcal{N} \!\left(\vecc^{(t)} \!\mid\! \overline{\vecm}^{(t)}, \overline{\bV}^{(t)} \right)$; here, the \emph{bar} on the variables denote the means and covariances of the \emph{approximated} Gaussian messages.
These approaches include the extended Kalman filter (EKF) (\cite{ekf2,ekf3,ekf}), which uses a linear approximation of the likelihood term around the point $\widetilde{\vecm}^{(t)}$, and thus reduce the non-Gaussian observation model to a Gaussian one; 
the unscented Kalman filter (UKF) (\cite{ukf2,ukf}), which uses the unscented transform (UT) to create a set of sigma vectors from $p(\vecc^{(t-1)})$ and uses them to approximate the mean and covariance of $\widehat{\alpha}(\vecc^{(t)})$ after the non-Gaussian observation;
and  Laplace approximations (\cite{laplace,gpml}), which use an iterative algorithm to find the mode of $\widehat{\alpha}(\vecc^{(t)})$ and the Hessian at the mode in order to approximate the mean and covariance of the approximated Gaussian messages. 
We will employ an approximation approach introduced in the expectation propagation (EP) literature (\cite{epminka}).

It is known that the specific values for $\overline{\vecm}^{(t)}$ and $\overline{\bV}^{(t)}$ that minimize the Kullback-Leibler (KL) divergence between $\mathcal{N}\!\left(\vecc^{(t)} \!\!\mid\!\! \overline{\vecm}^{(t)}, \overline{\bV}^{(t)} \right)$ and a target distribution $q(\vecc)$ are the first and second moments of $q(\vecc)$ \cite{gpml}. 
Luckily, for the probit observation model $p(Y^{(t)} \!\mid\! \vecc^{(t)}) = \! \Phi \! \left(  \left(2 Y^{(t)} -1 \right) \left( \vecw_{i^{(t)}}^T \vecc^{(t)}\! -\! \mu_{i^{(t)}} \right) \right)$, $\overline{\vecm}^{(t)}$, $\overline{\bV}^{(t)}$ and $b^{(t)}$ have closed-form expressions (see \fref{sec:appendix} for the details):
\begin{align}
\notag \overline{\vecm}^{(t)} & = \widetilde{\vecm}^{(t)} + \left(2 Y^{(t)} -1 \right) \frac{\widetilde{\bV}^{(t)} \vecw_{i^{(t)}}}{\sqrt{1+\vecw_{i^{(t)}}^T \widetilde{\bV}^{(t)} \vecw_{i^{(t)}}}} \frac{\mathcal{N}(z)}{\Phi(z)}, \\
\notag \overline{\bV}^{(t)} & = \widetilde{\bV}^{(t)} - \frac{\widetilde{\bV}^{(t)} \vecw_{i^{(t)}}\vecw_{i^{(t)}}^T \widetilde{\bV}^{(t)}}{1+\vecw_{i^{(t)}}^T \widetilde{\bV}^{(t)} \vecw_{i^{(t)}}} \left( z + \frac{\mathcal{N}(z)}{\Phi(z)} \right) \frac{\mathcal{N}(z)}{\Phi(z)}, \\
b^{(t)} & = \Phi(z), \label{eq:approx}
\end{align}
with 
\begin{align*}
z = \left(2 Y^{(t)} -1 \right) \frac{\vecw_{i^{(t)}}^T \widetilde{\vecm}^{(t)} - \mu_{i^{(t)}}}{\sqrt{1+\vecw_{i^{(t)}}^T \widetilde{\bV}^{(t)}\vecw_{i^{(t)}}}},
\end{align*}
and $\widetilde{\vecm}^{(t)}$ and $\widetilde{\bV}^{(t)}$ are given by \fref{eq:arecursion}.

SPARFA studied two different inverse link functions for analyzing binary-valued graded learner responses: the inverse probit link function and the inverse logit link function. In this application, the inverse probit link function is preferred over the inverse logit link function, due to the existence of the closed-form first and second moments described above. The inverse logit link function is not preferred as such convenient close-form expressions do not exist. Therefore, we will focus on the inverse probit link function in the sequel. 

Armed with the efficient approximation \fref{eq:approx}, the forward Kalman filtering message passing scheme described in \fref{sec:oldkf} can be applied to the problem at hand; 
the backward Kalman smoothing message passing scheme described in \fref{sec:oldks} remains unchanged. 
Using these recursions, estimates of the desired quantities $p(\vecc^{(t)} \!\mid\! \vecy^{(1)}, \ldots, \vecy^{(T)})$ can be computed efficiently, providing a way for learner concept knowledge tracing under the model \fref{eq:qa}.

\section{Content Analytics}
\label{sec:parameter}

So far, we have described an approximate Kalman filtering and smoothing approach for learner concept knowledge tracing, i.e., to estimate $p(\vecc_j^{(t)} \!\mid\! \vecy_j^{(1)}, \ldots, \vecy_j^{(T)}), \, \forall t,j$. 
The proposed method is only able to retrieve these estimates given both the observed binary graded learner responses $Y_j^{(t)}, \, \forall t,j$, and all learner initial knowledge parameters $\vecm_j^{(0)}, \bV_j^{(0)}, \, \forall j$, all learner concept knowledge state transition parameters $\bD_m$, $\vecd_m$, and $\boldsymbol{\Gamma}_m, \, \forall m$, and all question parameters, $\vecw_i$ and $\mu_i, \, \forall i$.

However, in a typical PLS, these parameters are unknown, in general, and need to be estimated from the observed data. 
Hence, we now detail a set of convex optimization-based techniques to estimate the parameters $\vecm_j^{(0)}, \bV_j^{(0)}, \, \forall j$, $\bD_m$, $\vecd_m$, and $\boldsymbol{\Gamma}_m, \, \forall m$, and $\vecw_i$, $\mu_i, \, \forall i$, given the estimates of the latent learner concept knowledge states $\vecc_j^{(t)}$ obtained from the approximate Kalman filtering approach described in \fref{sec:kf}.
Since the estimates of $\vecc_j^{(t)}$ are distributions rather than point estimates, SPARFA-Trace jointly traces learner concept knowledge and estimates learner, learning resource, and question-dependent parameters, using an expectation-maximization (EM) approach.

\subsection{SPARFA-Trace: An EM algorithm for parameter estimation}
\label{sec:em}

\fussy
EM has been widely used in the Kalman filtering framework to estimate the parameters of interest in the system (see \cite{kfbook} and \cite[Chap.~13]{bishop} for more details) due to numerous practical advantages (\cite{zoubinchapter}). 
SPARFA-Trace performs parameter estimation in an iterative fashion in the EM framework. 
All parameters are initialized to random initial values, and then, each iteration of the algorithm consist of two phases: (i) the current parameter estimates are used to estimate the latent state distributions $p(\vecc_j^{(t)} \!\mid\! \vecy_j^{(1)}, \ldots, \vecy_j^{(T)}), \, \forall t,j$; (ii), these latent state estimates are then used to maximize the expected joint log-likelihood of all the observed and latent state variables, i.e.,
\begin{align}
\label{eq:totalcost}
 & \underset{\vecm_j^{(0)}, \bV_j^{(0)}, \, \forall j, \bD_m, \vecd_m, \boldsymbol{\Gamma}_m, \forall m, \vecw_i, \mu_i, \, \forall i}{\text{maximize}} \quad \sum_{j=1}^N \mathbb{E}_{\vecc_j^{(1)}} \big[ \! \log \, p(\vecc_j^{(1)}\!\!\!\mid\! \vecm_j^{(0)}, \bV_j^{(0)}) \big]   +  \sum_{t=2}^T \sum_{j=1}^N \\ \notag & \mathbb{E}_{\vecc_j^{(t-1)},\vecc_j^{(t)}} \Big[ \!\! \log p(\vecc_j^{(t)} \!\mid\! \vecc_j^{(t-1)}, \bD_{m_j^{(t-1)}}, \vecd_{m_j^{(t-1)}}, \boldsymbol{\Gamma}_{m_j^{(t-1)}}) \Big] \!\!
 + \!\!\!\!\!\sum_{(t,j) \in \Omega_\text{obs}} \!\!\!\!\! \mathbb{E}_{\vecc_j^{(t)}} \!\big[ \! \log p(Y_j^{(t)} \!\mid\! \vecc_j^{(t)}, \vecw_{i_j^{(t)}}, \mu_{i_j^{(t)}})\big],
\end{align}
in order to obtain new (and hopefully improved) parameter estimates. 
SPARFA-Trace alternates between these two phases until convergence, i.e., a maximum number of iterations is reached or the change in the estimated parameters between two consecutive iterations falls below a given threshold. 
\sloppy

\subsection{Estimating the learner initial knowledge parameters}
\label{sec:learnerparam}

We start with the estimation method for the learner initial knowledge parameters $\vecm_j^{(0)}, \bV_j^{(0)}, \, \forall j$. To this end, we minimize the expected negative log-likelihood for the $j^\text{th}$ learner:
\begin{align*}
& \mathbb{E}_{\vecc_j^{(1)}} \!\big[ - \log \, p(\vecc_j^{(1)} \!\mid\! \vecm_j^{(0)}, \bV_j^{(0)}) \big] 
 = \frac{1}{2} \log | \bV_j^{(0)} | +  \mathbb{E}_{\vecc_j^{(1)}} \!\!\left[ \frac{1}{2} (\vecc_j^{(1)} - \vecm_j^{(0)})^T (\bV_j^{(0)})^{-1} (\vecc_j^{(1)} - \vecm_j^{(0)}) \right]\!,
\end{align*}
where $|\bV_j^{(0)}|$ denotes the determinant of the covariance matrix $\bV_j^{(0)}$.
Since we do not impose constraints on $\vecm_j^{(0)}$ and $\bV_j^{(0)}$, these estimates can be obtained as
\begin{align*}
\vecm_j^{(0)} & = \mathbb{E}_{\vecc_j^{(1)}} \!\left[ \vecc_j^{(1)} \right] = \widehat{\vecm}_j^{(1)} \quad \text{and} \quad
\bV_j^{(0)}  = \mathbb{E}_{\vecc_j^{(1)}} \!\left[ (\vecc_j^{(1)}-\widehat{\vecm}_j^{(1)}) (\vecc_j^{(1)} - \widehat{\vecm}_j^{(1)})^T \right] = \widehat{\bV}_j^{(1)},
\end{align*}
where the estimates $\widehat{\vecm}_j^{(1)}$ and $\widehat{\bV}_j^{(1)}$ are obtained from the Kalman smoothing recursions  \fref{eq:bp} in \fref{sec:oldks}.

\subsection{Estimating the learner concept knowledge state transition parameters}
\label{sec:docparam}

Now we estimate the latent learner concept knowledge state transition (i.e., learning resource) parameters $\bD_m$, $\vecd_m$, and $\boldsymbol{\Gamma}_m, \, \forall m$. To this end, define $\mathcal{M}^m$ as the set containing time and learner indices ($t,j$), indicating that learner $j$ studies the $m^\text{th}$ learning resource between time instances $t-1$ and $t$. 
With this definition, we aim to minimize the expected negative log-likelihood 
\begin{align*}
& \sum_{t,j: (t,j) \in \mathcal{M}^m} \mathbb{E}_{\vecc_j^{(t-1)},\vecc_j^{(t)}} [ - \log p(\vecc_j^{(t)} \!\mid\! \vecc_j^{(t-1)}, \bD_m, \vecd_m, \boldsymbol{\Gamma}_m) ]\\
& = \sum_{t,j: (t,j) \in \mathcal{M}^m} \Bigg( \frac{1}{2} \log |\boldsymbol{\Gamma}_m| \\
& + \mathbb{E}_{\vecc_j^{(t-1)},\vecc_j^{(t)}} \! \left[ \frac{1}{2} (\vecc_j^{(t)} - \vecc_j^{(t-1)} - \bD_m \vecc_j^{(t-1)} - \vecd_m)^T \boldsymbol{\Gamma}_m^{-1} (\vecc_j^{(t)} - \vecc_j^{(t-1)} - \bD_m \vecc_j^{(t-1)} - \vecd_m) \right] \!\Bigg) 
\end{align*}
subject to the assumptions (A4)--(A6). We start by estimating $\bD_m$ and $\vecd_m$ given $\boldsymbol{\Gamma}_m$, and then use these estimates to estimate $\boldsymbol{\Gamma}_m$. In order to induce sparsity on $\bD_m$ to take (A6) into account, we impose an $\ellone$-norm penalty on $\bD_m$ (\cite{tibsbook}). Taking only the terms containing $\bD_m$ and $\vecd_m$, we can formulate the following augmented optimization problem:
\begin{align*}
\text{(P}_\text{d}\text{)} \quad \quad \underset{\bD_m \in \mathcal{L}^+, \vecd_m}{\text{minimize}} \sum_{t,j: (t,j) \in \mathcal{M}^m} \mathbb{E}_{\vecc_j^{(t-1)},\vecc_j^{(t)}} \Big[ (\widetilde{\bD}_m \widetilde{\vecc}_j^{(t-1)})^T \boldsymbol{\Gamma}_m^{-1} (\widetilde{\bD}_m \widetilde{\vecc}_j^{(t-1)}) - & \\
 (\vecc_j^{(t)} - \vecc_j^{(t-1)})^T \boldsymbol{\Gamma}_m^{-1} (\vecc_j^{(t)} - \vecc_j^{(t-1)}) \Big] + \gamma\normone{\bD_m},& 
\end{align*}
where $\mathcal{L}^+$ denotes the set of lower-triangular matrices with non-negative entries.
For notational simplicity, we have written $[\bD_m \; \vecd_m]$ as $\widetilde{\bD}_m$. We also write the augmented latent state vectors $[(\vecc_j^{(t-1)})^T \; 1]^T$ as $\widetilde{\vecc}_j^{(t-1)}$, when multiplied by $\widetilde{\bD}_m$, correspondingly. Note that the $\ellone$-norm penalty only applies to the matrix $\bD_m$ in the used notation. 

The problem (P$_\text{d}$) is convex in $\widetilde{\bD}_m$, and hence, can be solved efficiently. In particular, we use the fast iterative shrinkage and thresholding algorithm (FISTA) framework (\cite{fista}). The FISTA algorithm starts with a random initialization of $\widetilde{\bD}_m$ and iteratively updates $\widetilde{\bD}_m$ until a maximum number of iterations $\ell_\text{max}$ is reached or the change in the estimate of $\widetilde{\bD}_m$ between two consecutive iterations falls below a certain threshold. In each iteration $\ell = 1,2,\ldots, \ell_\text{max}$, the algorithm performs two steps. First, a gradient step that aims to lower the cost function performs
\begin{align}
\label{eq:docgrad}
\widehat{\bD}_m^{\ell+1} \leftarrow \widetilde{\bD}_m^\ell - \eta_\ell \nabla f(\widetilde{\bD}_m),
\end{align}
where $f(\widetilde{\bD}_m)$ corresponds to the differentiable part of the cost function (excluding the $\ellone$-norm penalty) in (P$_\text{d}$). The quantity $\eta_\ell$ is a step size parameter for iteration $\ell$. For simplicity, we will take $\eta_\ell = 1/L$ in all iterations, where $L$ is the Lipschitz constant given by
\begin{align*}
L = \sigma_\text{max}\Bigg(\sum_{t,j: (t,j) \in \mathcal{M}^m} \mathbb{E}_{\vecc_j^{(t-1)},\vecc_j^{(t)}}[(\vecc_j^{(t)} - \vecc_j^{(t-1)}) (\vecc_j^{(t-1)})^T]\Bigg) \; \sigma_\text{max}(|\mathcal{M}^m| \,\boldsymbol{\Gamma}_m^{-1}).
\end{align*}
Here $\sigma_\text{max}(\cdot)$ denotes the maximum singular value of a matrix, and $|\mathcal{M}^m|$ denotes the cardinality of the set $\mathcal{M}^m$. The gradient $\nabla f(\widetilde{\bD}_m)$ in \fref{eq:docgrad} is given by
\begin{align*}
\nabla \! f(\widetilde{\bD}_m) & \! = \!  - \boldsymbol{\Gamma}_m^{-1} \! \sum_{t,j: (t,j) \in \mathcal{M}^m} \! \big(\mathbb{E}_{\vecc_j^{(t-1)},\vecc_j^{(t)}}[(\vecc_j^{(t)} - \vecc_j^{(t-1)}) (\widetilde{\vecc}_j^{(t-1)})^T] \! -\! \bD_m^\ell \mathbb{E}_{\vecc_j^{(t-1)}} [\widetilde{\vecc}_j^{(t-1)} (\widetilde{\vecc}_j^{(t-1)})^T] \big) \\
& = \!  - \boldsymbol{\Gamma}_m^{-1} \! \sum_{t,j: (t,j) \in \mathcal{M}^m} \Bigg([ \bJ_j^{(t-1)} \widehat{\bV}_j^{(t)} \! + \! \widehat{\vecm}_j^{(t)} (\widehat{\vecm}_j^{(t-1)})^T \! - \! \widehat{\bV}_j^{(t-1)} \! -\! \widehat{\vecm}_j^{(t-1)} (\widehat{\vecm}_j^{(t-1)})^T \\ 
& \quad \widehat{\vecm}_j^{(t)} - \widehat{\vecm}_j^{(t-1)} ] - \bD_m^\ell \left[ \begin{array}{cc} \widehat{\bV}_j^{(t-1)} + \widehat{\vecm}_j^{(t-1)} (\widehat{\vecm}_j^{(t-1)})^T & \widehat{\vecm}_j^{(t-1)} \\ (\widehat{\vecm}_j^{(t-1)})^T & 1 \end{array}\right] \Bigg).
\end{align*}
The parameters $\bJ_j^{(t-1)}$, $\widehat{\vecm}_j^{(t-1)}$, $\widehat{\vecm}_j^{(t)}$, $\widehat{\bV}_j^{(t-1)}$, and $\widehat{\bV}_j^{(t)}$ are obtained from the backward recursions in \fref{eq:bp}.
Next, the FISTA algorithm performs a projection step, which takes into account the sparsifying regularizer $\gamma \normone{\bD_m}$, and the assumptions (A4) and (A5):
\begin{align}
\label{eq:docproj}
\widetilde{\bD}_m^{\ell+1} \leftarrow P_{\mathcal{L}^+}( \text{max} \{\widehat{\bD}_m^{\ell+1} - \gamma \eta_\ell, 0\}),
\end{align}
where $P_{\mathcal{L}^+}(\cdot)$ correspond to the projection onto the set of lower-triangular matrices by setting all entries in the upper triangular part of $\bD_m^{\ell+1}$ to zero. The maximum operator operates element-wise on $\bD_m^{\ell+1}$. 
The updates \fref{eq:docgrad} and \fref{eq:docproj} are repeated until convergence, eventually providing a new estimate $\widetilde{\bD}_m^\text{new}$ for $[\bD_m \; \vecd_m]$. 

Using these new estimates, the update for $\boldsymbol{\Gamma}_m$ can be computed in closed form:
\begin{align*}
\boldsymbol{\Gamma}_m^\text{new} & = \frac{1}{|\mathcal{M}^m|} \sum_{t,j: (t,j) \in \mathcal{M}^m} \big(\mathbb{E}_{\vecc_j^{(t)}} [\vecc_j^{(t)} (\vecc_j^{(t)})^T] - \widetilde{\bD}_m^\text{new} \mathbb{E}_{\vecc_j^{(t-1)},\vecc_j^{(t)}}[\widetilde{\vecc}_j^{(t-1)} (\vecc_j^{(t)})^T]  \\
& \quad - \mathbb{E}_{\vecc_j^{(t-1)},\vecc_j^{(t)}}[\vecc_j^{(t)} (\widetilde{\vecc}_j^{(t-1)})^T] (\widetilde{\bD}_m^\text{new})^T + (\widetilde{\bD}_m^\text{new})\mathbb{E}_{\vecc_j^{(t-1)}}[\widetilde{\vecc}_j^{(t-1)} (\widetilde{\vecc}_j^{(t-1)})^T](\widetilde{\bD}_m^\text{new})^T\big) \\
& = \frac{1}{|\mathcal{M}^m|} \sum_{t,j: (t,j) \in \mathcal{M}^m} \Big(\widehat{\bV}_j^{(t)} + \widehat{\vecm}_j^{(t)} (\widehat{\vecm}_j^{(t)})^T - \widetilde{\bD}_m^\text{new} \left[ \begin{array}{c}\bJ_{t-1,j} \widehat{\bV}_j^{(t)} + \! \widehat{\vecm}_j^{(t)} (\widehat{\vecm}_j^{(t-1)})^T \\ (\widehat{\vecm}_j^{(t)})^T \end{array} \right]\\
& \quad - \big[ \bJ_j^{(t-1)} \widehat{\bV}_j^{(t)} + \! \widehat{\vecm}_j^{(t)} (\widehat{\vecm}_j^{(t-1)})^T \;\; \widehat{\vecm}_j^{(t)} \big]  (\widetilde{\bD}_m^\text{new})^T \\
& \quad + \widetilde{\bD}_m^\text{new} \left[ \begin{array}{cc} \widehat{\bV}_j^{(t-1)} + \widehat{\vecm}_j^{(t-1)} (\widehat{\vecm}_j^{(t-1)})^T & \widehat{\vecm}_j^{(t-1)} \\ (\widehat{\vecm}_j^{(t-1)})^T & 1 \end{array}\right](\widetilde{\bD}_m^\text{new})^T\Big).
\end{align*}

\subsection{Estimating the question-dependent parameters}
\label{sec:qparam}

\sloppy
We next show how to estimate the question-dependent parameters $\vecw_i$, $\mu_i, \, \forall i$. To this end, we define $\mathcal{Q}^i$ as the collection set of time and learner indices ($t,j$) that learner $j$ answered the $i^\text{th}$ question at time instance $t$. We then minimize the expected negative log-likelihood of all the observed binary-valued graded learner responses \fref{eq:qa} for the $i^{th}$ question, subject to assumptions (A2) and (A3) on the question--concept association vector~$\vecw_i$. 
\fussy
In order to impose sparsity on $\vecw_i$, we add an $\ellone$-norm penalty to the cost function, which leads to the following optimization problem:
\begin{align*}
(\text{P}_\text{w}) \quad \quad \underset{\vecw_i: w_{i,k} \geq 0, \forall k}{\text{minimize}}  \sum_{(t,j) \in \mathcal{Q}^i} \mathbb{E}_{\vecc_j^{(t)}} \!\left[- \text{log} \,\Phi( (2Y_j^{(t)}-1) (\vecw_i^T \vecc_j^{(t)} - \mu_i)) \right] + \lambda \normone{\vecw_i}.
\end{align*}
This problem corresponds to the (RR$_1^+$) problem of SPARFA detailed in \cite{sparfa}, where point estimates of~$\vecc_j$ are given and the problem is convex in $\vecw_i$. 
In particular, given the distribution $\vecc_j^{(t)} \distas \mathcal{N} (\vecc_j^{(t)} \!\mid\! \widehat{\vecm}_j^{(t)}, \widehat{\bV}_j^{(t)})$, (P$_\text{w}$) is still convex in $\vecw_i$, thanks to the linearity of the expectation operator.
 However, the inverse probit link function prohibits us from obtaining a simple form of this expectation. 
In order to develop a tractable algorithm to approximately solve this problem, we utilize the unscented transform (UT) (\cite{ukf}) to approximate the cost function of (P$_\text{w}$). 

The UT is commonly used in Kalman filtering literature to approximate the statistics of a random variable undergoing non-linear transformations. Specifically, given a $K$-dimensional random variable $\vecx$ with known mean and covariance and a non-linear function $g(\cdot)$, the UT generates a set of $2K+1$ so-called \emph{sigma vectors} $\{\mathcal{X}_n\}$ and a set of corresponding weights $\{u_n\}$ as detailed in \cite[Eq.15]{ukf}, in order to approximate the mean and covariance of the vector $\vecy = g(\vecx)$. 
As shown in \cite{ukf}, this approximation is accurate up to the third order for Gaussian distributed random variables $\vecx$.

Following the paradigms of the UT, we generate a set of sigma vectors $\{(\widetilde{\vecc}_j^{(t)})_n\}$ and a corresponding set of weights $\{u_n\}$, $n \in \{1, \ldots, 2K+1\}$, for each latent state vector $\vecc_j^{(t)}$, given the mean $\widehat{\vecm}_j^{(t)}$ and covariance $\widehat{\bV}_j^{(t)}$. For computational simplicity, we will use the same set of weights for all latent state vectors $\vecc_j^{(t)}$. The optimization problem (P$_\text{w}$) can now be approximated by
\begin{align*}
\underset{\vecw_i: w_{i,k} \geq 0, \forall k}{\text{minimize}} \sum_{(t,j) \in \mathcal{Q}^i}  \sum_{n=1}^{2K+1} u_n \left(- \text{log} \,\Phi( (2Y_j^{(t)}-1) (\vecw_i^T (\widetilde{\vecc}_j^{(t)})_n - \mu_i)) \right) + \lambda \normone{\vecw_i},
\end{align*}
which, once again, can be solved efficiently by using the FISTA framework. The resulting iterative procedure performs two steps in each iteration $\ell$: First, a gradient step that aims at lowing the cost function performs 
\begin{align} 
\label{eq:probgrad}
\widehat{\vecw}_i^{\ell+1} \gets \vecw_i^\ell - \eta_\ell \, \nabla f(\vecw_i),
\end{align}
where $f(\vecw_i)$ corresponds to the differentiable portion (excluding the $\ellone$-norm penalty part) of the cost function in (P$_\text{w}$). The gradient $\nabla f(\vecw_i)$ is given by
$\nabla f(\vecw_i) = - \widetilde{\bC_i} \widetilde{\vecr}_i$,
where $\widetilde{\vecr}_i$ is a $(2K+1) |\mathcal{Q}^i| \times 1$ vector $ \vecr_i = [\veca_i^{1} \ldots, \veca_i^{|\mathcal{Q}^i|}]^T$.
The vector $\veca_i^{q}$ is defined by $\veca_i^{q} =\left[(g_i^q)_1, \ldots, (g_i^q)_{2K+1} \right]$, where
\begin{align*}
(g_i^q)_n = u_n 2(Y_{j_q}^{(t_q)}-1)  \frac{\mathcal{N}\left(2(Y_{j_q}^{(t_q)}-1)\vecw_i^T (\widetilde{\vecc}_{j_q}^{(t_q)})_n\right)}{\Phi\left(2(Y_{j_q}^{(t_q)}-1)\vecw_i^T (\widetilde{\vecc}_{j_q}^{(t_q)})_n\right)}, 
\end{align*}
in which $(t_q,j_q)$ represents the $q^\text{th}$ time--learner index pair in $\mathcal{Q}^i$. The $K \times (2K+1) |\mathcal{Q}^i|$ matrix $\widetilde{\bC}_i$ is defined as $\widetilde{\bC}_i = \left[ (\bG_i)_1, \ldots, (\bG_i)_{|\mathcal{Q}^i|} \right]$, where the $K \times (2K+1)$ matrix $(\bG_i)_q$ is given by
\begin{align*}
(\bG_i)_q = \left[(\widetilde{\vecc}_{j_q}^{(t_q)})_1, \; \ldots, \; (\widetilde{\vecc}_{j_q}^{(t_q)})_{2K+1}\right]. 
\end{align*}
The quantity $\eta_\ell$ is a step size parameter for iteration $\ell$. For simplicity, we will take $\eta_\ell = 1/L$ in all iterations, where $L$ is the Lipschitz constant given by
$L = \sigma_\text{max}(\widetilde{\bC}_i) \, \sigma_\text{max}(\widetilde{\bC}_i')$,
where $\widetilde{\bC}_i'$ is a $K \times (2K+1) |\mathcal{Q}^i|$ matrix defined as $\widetilde{\bC}_i' = \left[ (\bG_i')_1, \ldots, (\bG_i')_{|\mathcal{Q}^i|} \right]$, where the $K \times (2K+1)$ matrix $(\bG_i')_q$ is given by
\begin{align*}
(\bG_i')_q = \left[ u_1 (\widetilde{\vecc}_{j_q}^{(t_q)})_1, \; \ldots, \; u_{2K+1} (\widetilde{\vecc}_{j_q}^{(t_q)})_{2K+1} \right]. 
\end{align*}

Next, the FISTA algorithm performs a projection step, which takes into account $\lambda \normone{\vecw_i}$ and the assumption (A3):
\begin{align}\label{eq:probproj} 
\vecw_i^{\ell+1} \gets \max\{\widehat{\vecw}_i^{\ell+1}-\lambda \eta_\ell,0\}.
\end{align}

The steps \fref{eq:probgrad} and \fref{eq:probproj} are repeated until convergence, providing a new estimate $\vecw_i^\text{new}$ of the question--concept association vector $\vecw_i$. For simplicity of exposition, the question intrinsic difficulties $\mu_i$ are omitted in the derivations above, as they can be included as an additional entry in $\vecw_i$ as $[\vecw_i^T \; \mu_i]^T $; the corresponding latent learner concept knowledge state vectors $\vecc_j^{(t)}$ are augmented as $[(\vecc_j^{(t)})^T \; 1]^T$.

\section{Experimental Results}
\label{sec:sims}

We now demonstrate the efficacy of SPARFA-Trace on synthetic and real-world educational datasets. We start by performing experiments using synthetic data to demonstrate that SPARFA-Trace is able to accurately trace latent learner concept knowledge and accurately estimate learner concept knowledge state transition parameters and question-dependent parameters. We then compare SPARFA-Trace against two established methods on predicting unobserved binary-valued learner response data, namely knowledge tracing (KT) (\cite{kt,ktcomparepardos}) and SPARFA (\cite{sparfa}). Finally, we show how SPARFA-Trace is able to visualize learners' concept knowledge state evolution over time, and the learning resource and question quality and their content organization. For all the synthetic and real data experiments shown next, the regularization parameters $\lambda$, $\gamma$ and $\sigma_0^2$ are chosen via cross-validation (\cite{tibsbook}), and all experiments are repeated for $25$ independent Monte--Carlo trials for each instance of the model parameter we control.

\subsection{Experiments with synthetic data}
\label{sec:synth}

In the following experiments with synthetic data, we assess the performance of SPARFA-Trace in both (i) learner concept knowledge tracing, and (ii) estimating all learner concept knowledge state transition parameters and question-dependent parameters. 

\sloppy
\paragraph{Dataset:} We generate the learning resource-induced learner knowledge state transition parameters $\bD_m$, $\vecd_m$, $\boldsymbol{\Gamma}_m$, $m \in \{1,\ldots, M\}$, $\vecw_i, \mu_i, i \in \{1,\ldots, Q\}$, under the assumptions (A1)--(A6), and randomly generate learner prior parameters $\vecm_j^{(0)}$ and $\bV_j^{(0)}, j \in \{1,\ldots,N \}$. 
Using these parameters, we randomly generate latent learner concept knowledge states $\vecc_j^{(t)}$ and observed binary-valued graded responses $Y_j^{(t)}$, $t \in \{1,\ldots,T\}$, according to \fref{eq:qa} and \fref{eq:trans}. 
The number of time instances is $T=100$, and one question is assigned to every learner at every time instance, so $Q=T=100$. 
The dataset consist of $10$ assignment sets, each consisting of $10$ questions. 
The learners' concept knowledge states evolve between consecutive assignment sets, induced by their interaction with learning resources. Therefore, the number of learning resources is $M=9$. 
There are a total of $K=5$ concepts, as this choice is shown to be reasonable for real-world educational scenarios (see, e.g., \cite{nonparamsparfab} for a corresponding discussion).
\fussy

\sloppy
\paragraph{Learner concept knowledge tracing:} For the learner concept knowledge state estimation experiment, we fix the number of learners as $N=50$ and vary the percentage of observed entries in the $Q \times N$ learner response matrix $\bY$ as $\{100\%, 75\%, 50\%, 25\%\}$ and calculate the normalized concept knowledge state estimation error
\begin{align} \label{eq:errmetric}
E_\vecc = \frac{1}{NT}\sum_{(t,j)}\frac{\|\vecm_j^{(t)}-\vecc_j^{(t)}\|^2_2}{\|\vecc_j^{(t)}\|^2_2}. 
\end{align}
\fussy
In this experiment, all learner-dependent and learner concept knowledge state transition and question parameters are assumed to be known. Thus, we only run the Kalman filtering and smoothing part of SPARFA-Trace. 

\begin{figure*}[t]
\centering
\hspace{-.5cm}
\subfigure[]{
\includegraphics[width=0.45\textwidth]{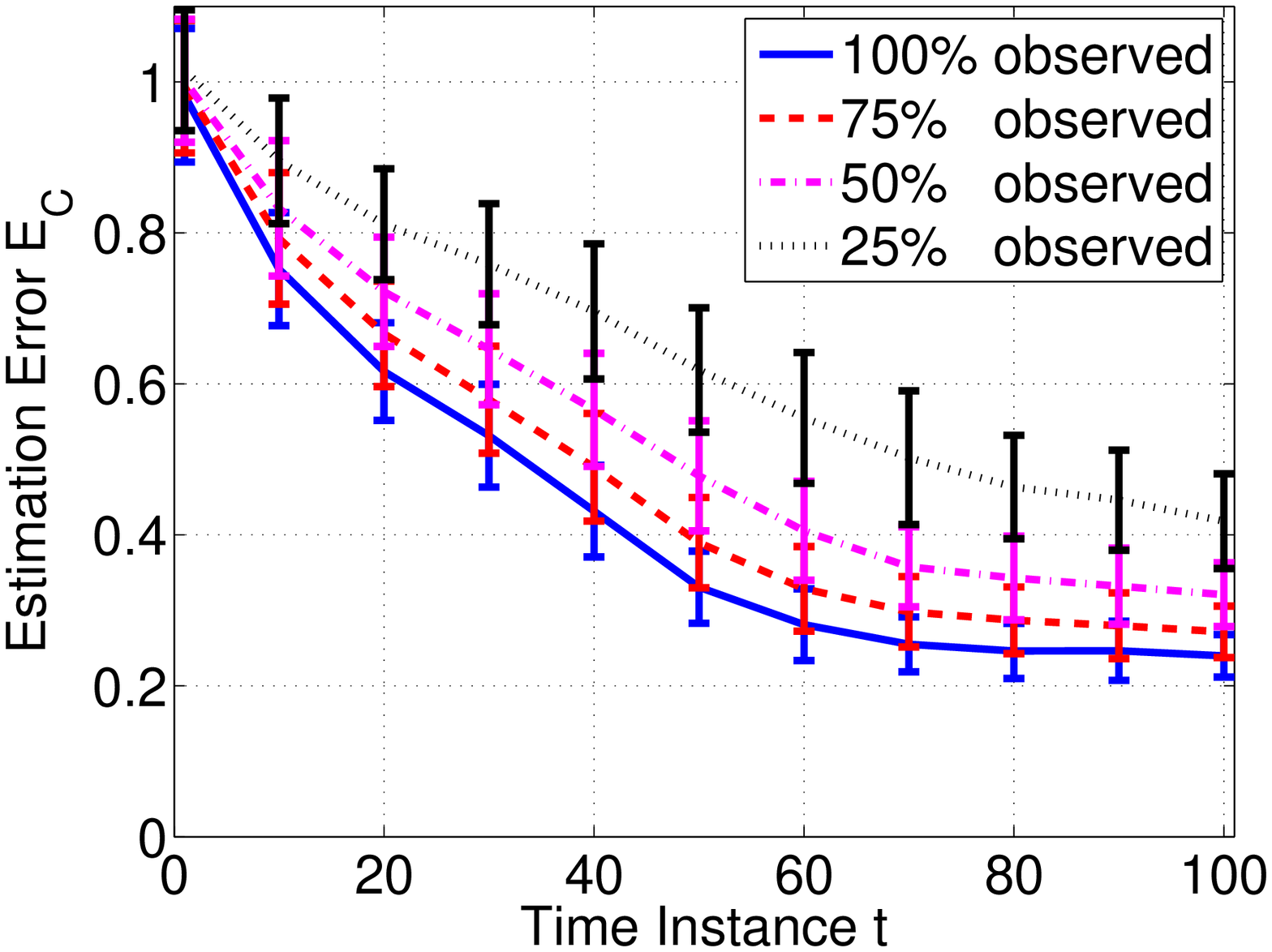}
\label{fig:synth_1}
}
\hspace{.5cm}
\subfigure[]{
\includegraphics[width=0.45\textwidth]{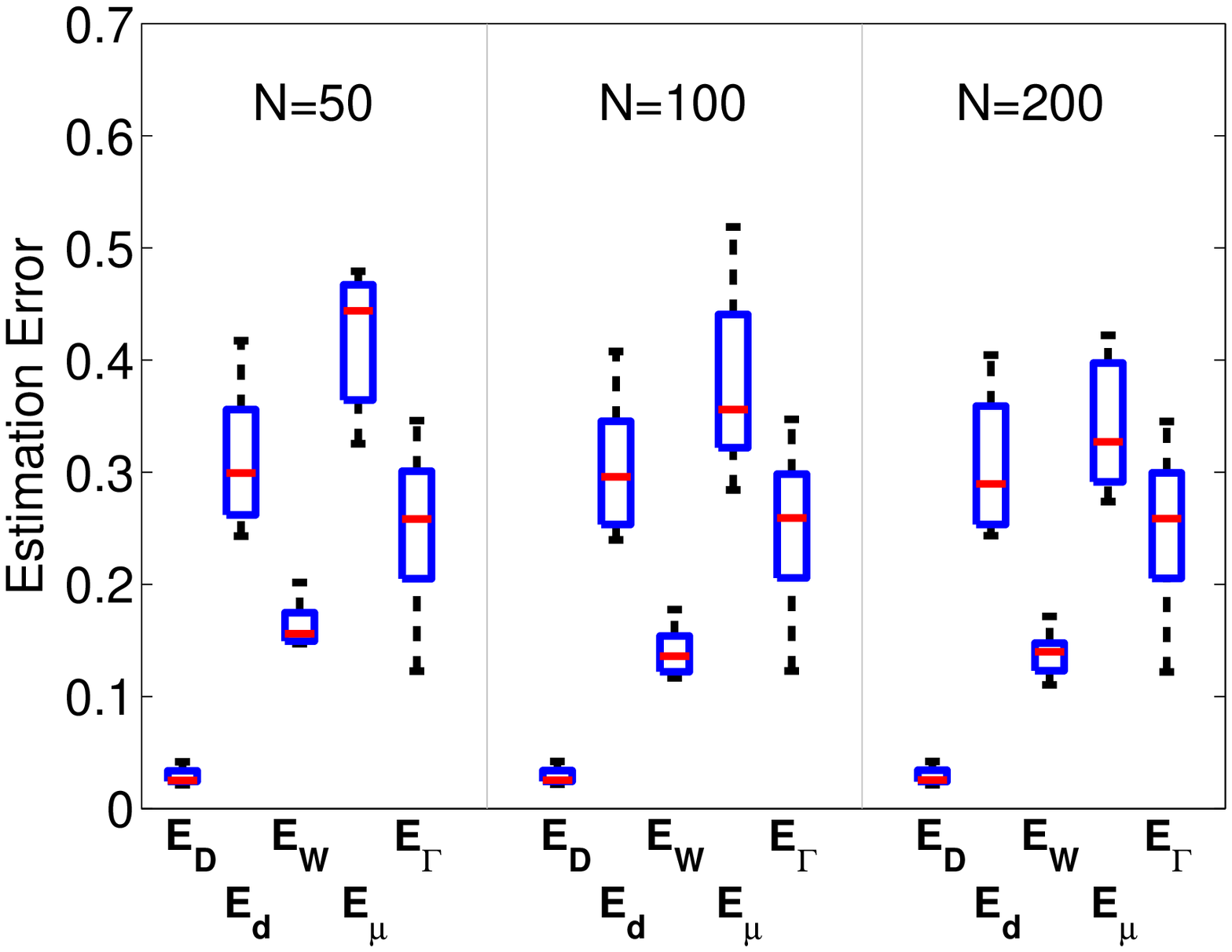}
\label{fig:synth_2}
}\\[0.3cm]
\vspace{-0.3cm}
\caption{Accuracy of latent concept knowledge state and learning resource and question-dependent parameters estimation for synthetic data. (a) Learner concept knowledge state estimation error versus time instance $t$ for different percentages of observed responses; (b) Learning resource parameter estimation error for various number of learners $N$. Note the general trend that all considered performance measures improve as the amount of observed data  increases.}
\label{fig:synth}
\end{figure*}

\fref{fig:synth_1} shows the results from the learner concept knowledge state estimation experiment. We observe that the estimation of learner concept knowledge states becomes increasingly accurate as time proceeds. The performance of SPARFA-Trace decreases as the percentage of missing observations increases. Moreover, SPARFA-Trace can still obtain accurate estimates of $\vecc_j^{(t)}$ even when only a small portion of the response data is observed.

\paragraph{Estimating learner concept knowledge state transition and question parameters:}
To assess SPARFA-Trace on the estimation performance of learner concept knowledge state transition and question parameters, we perform a second experiment, which focus on the estimation of all learning resource and question-dependent parameters: $\bD_m, \vecd_m, \boldsymbol{\Gamma}_m, \, \forall m$, $\vecw_i, \mu_i,\, \forall i$. The learner concept knowledge states $\vecc_j^{(t)}$ are not given and are estimated simultaneously, while we treat the learner prior parameters $\vecm_j^{(0)}$ and $\bV_j^{(0)}, j \in \{1,\ldots,N \}$ as given, to avoid the scaling unidentifiability issue in the model (one can arbitrarily scale the learner concept knowledge state vectors $\vecc_j^{(t)}$ and adjust the scale of the question--concept association vectors $\vecw_i$ accordingly, and still arrive at the same likelihood for the observations. See, e.g., \cite{sparfa} for a detailed discussion.) We fix the number of concepts as $K=5$, vary the number of learners as $N \in \{ 50, 100, 200\}$, and examine the estimation error of SPARFA-Trace on all instructional and question-dependent parameters using a similar metric as in \fref{eq:errmetric}. The observed learner response matrix $\bY$ is assumed to be fully observed. We run SPARFA-Trace until convergence, to provide estimates of all unknown parameters.

\fref{fig:synth_2} shows the box-and-whisker plots of the estimation error on all five types of parameters for different numbers of learner $N$. We can see that the parameter estimation performance of SPARFA-Trace improves as the number of learners increase. More importantly, SPARFA-trace provides accurate estimates of these parameters even when the problem size is relatively small (e.g., the number of learners $N=50$). 

In summary of these synthetic experiments, we can conclude that SPARFA-Trace is capable of accurately estimating both latent learner concept knowledge states and the learner concept knowledge state transition and question parameters.

\subsection{Predicting responses for new learners}
\label{sec:ktcompare}

We now compare SPARFA-Trace against the KT method described in \cite{ktcomparepardos} on predicting responses for new learners that do not have previous recorded response history. 

\paragraph{Dataset:} The dataset we use for this experiment is from an undergraduate computer engineering course collected from OpenStax Tutor (OST) (\cite{ost}). We will refer to this dataset as ``Dataset~1'' in the following experiments. This dataset consist of the binary-valued graded response from $92$ learners answering $203$ questions, with $99.5\%$ of the responses observed. Since the KT implementation of \cite{ktcomparepardos} is unable to handle missing data, we removed learners that do not answer every question from the dataset, resulting in a pruned dataset of $73$ learners. The course is organized into three independent sections: The first section is on digital logic, the second on data structures, and the third on basic programming concepts. The full course consist of $11$ assessments, including $8$ homework assignments and an exam at the end of each section; we assume that the learners' concept knowledge state transitions can only happen between two consecutive assignments/exams, due to their interaction with all the lectures/readings/exercises.

\paragraph{Experimental setup:}
Since KT is only capable of handling educational datasets that involve a single concept, we partition Dataset~1 into three parts, with each part corresponding to one of the three independent sections. We run KT independently on the three parts, and aggregate the prediction results. (We also ran KT on the entire Dataset~1 without separating it into 3 independent sections. The obtained results are inferior to those obtained by running KT on 3 independent sections.)  The four parameters of KT (learner prior, learning probability, guessing probability, slipping probability) are initialized with the best initial value we find over 5 different initializations. For SPARFA-Trace, we use $K=3$, with each concept corresponding to one section of the dataset. In order to alleviate the identifiability issue in our model, we initialize the algorithm with $\vecw_{i,k} = 1$ where question $i$ is in section $k$ and $\vecw_{i,k} = 0$ otherwise. We also initialize the matrices $\bD_m$ with identity matrices $\bI_{3 \times 3}$, the vectors $\vecd_m$ with zero vectors, and covariance matrices $\boldsymbol{\Gamma}_m$ with identity matrices. 

For cross-validation, we randomly partition Dataset~1 into $5$ folds, with each fold consisting of $1/5$ of the learners answering all questions. Four folds of the data are used as the training set and the other fold is used as the test set. We train both KT and SPARFA-Trace on the training set and obtain estimates on all learner, learning resource and question-dependent parameters, and test their prediction performances on the test set. For previously unobserved new learners in the test set, both algorithms make the first prediction of $Y_j^{(1)}$ at $t=1$ using question-dependent parameters estimated from the training set. As time goes on, more and more observed responses $Y_j^{(t)}$ are available to both algorithms, and they use these responses to make future predictions.

We compare both algorithms on three metrics: prediction accuracy, prediction likelihood, and area under the receiver operation characteristic (ROC) curve. The prediction accuracy corresponds to the percentage of correctly predicted responses; the prediction likelihood corresponds to the average the predicted likelihood of the unobserved responses, i.e., $\frac{1}{\mid \Omega_\text{obs}^c \mid} \textstyle \sum_{t,j:(t,j)\in \Omega_\text{obs}^c} p(Y_j^{(t)} | \vecw_{i_j^{(t)}},\vecc_j^{(t)})$ where $\Omega_\text{obs}^c$ is the set of learner responses in the test set; the area under the ROC curve is a commonly-used performance metric for binary classifiers (see \cite{ktpardos} for details). The area under the ROC curve always is always between 0 and 1, with a larger value representing higher classification accuracy.

Since SPARFA-Trace does not provide point estimates of $\vecc_j^{(t)}$ but rather their distributions, we compute the predicted likelihood of unobserved responses by: 
\begin{align*}
\mathbb{E}_{\vecc_j^{(t)}} \left[ p(Y_j^{(t)} | \vecw_{i_j^{(t)}},\vecc_j^{(t)}) \right]=  \Phi \left( \left(2Y_j^{(t)}-1\right) \frac{\vecw_{i_j^{(t)}}^T \widehat{\vecm}_j^{(t)} - \mu_{i_j^{(t)}}}{\sqrt{1 + \vecw_{i_j^{(t)}}^T \widehat{\bV}_j^{(t)} \vecw_{i_j^{(t)}}}} \right).
\end{align*}

\paragraph{Results:}
The means and standard deviations of all three metrics covering multiple cross-validation trials are shown in \fref{tbl:ktcompare}. We can see that SPARFA-Trace outperforms KT on all performance metrics for Dataset~1. We also emphasize that SPARFA-Trace is capable of achieving superior prediction performance while simultaneously estimating the quality and content organization parameters of all learning resources and questions. 
\begin{table}
\centering
\caption{Comparisons of SPARFA-Trace against knowledge tracing (KT) on predicting responses for new learners using using Dataset~1. SPARFA-Trace outperforms KT on all considered metrics.}
\label{tbl:ktcompare}
\vspace{0.3cm}
\begin{tabular}{lll}
\toprule
~&KT& {\bf SPARFA-Trace} \\
\midrule
Prediction accuracy \quad \quad & $86.42 \pm 0.16 \%$ \quad \quad & \boldsymbol{$87.49 \pm 0.12 \%$}  \\ 
Prediction likelihood \quad \quad & $0.7718 \pm 0.0011$ \quad \quad & \boldsymbol{$0.8128 \pm 0.0044$}  \\ 
Area under the ROC curve \quad \quad & $0.5989 \pm 0.0056$ \quad \quad & \boldsymbol{$0.8157 \pm 0.0028$}  \\
\bottomrule
\end{tabular}
\end{table}

\subsection{Predicting unobserved learner responses}
\label{sec:sparfacompare}

It has been shown (\cite{pfa,ktpardos}) that collaborative filtering methods often outperform KT in predicting unobserved learner responses, even though they ignore any temporal evolution aspects of the dataset. Hence, we compare SPARFA-Trace against the original SPARFA framework (\cite{sparfa}), which shows state-of-the-art collaborative filtering performance on predicting unobserved learner responses. 

\paragraph{Datasets:} 
We will use two datasets in this experiment. The first dataset is the full Dataset~1 with $92$ learners answering $203$ questions, explained in \fref{sec:ktcompare}. The second dataset we use is from a signals and systems undergraduate course on OST, consisting of $41$ learners answering $143$ questions, with $97.1\%$ of the responses observed. We will refer to this dataset as ``Dataset~2'' in the following experiments. All the questions were manually labeled with a number of $K=4$ concepts, with the concepts being listed in \fref{fig:Wex_b}. The full course consist of $14$ assessments, including $12$ assignments and $2$ exams; we will treat all the lectures/readings/exercises the learners interact with between two consecutive assignments/exams as an learning resource.

\paragraph{Experimental setup:} 
We randomly partition the $143 \times 43$ (or $203 \times 92$) matrix $\bY$ of observed graded learner responses into $5$ folds for cross-validation. Four folds of the data are used as the training set and the other fold is used as the test set. We train both the probit variant of SPARFA-M and SPARFA-Trace on the training set to estimate the learner concept knowledge states and the learner, learning resource and question-dependent parameters, and then use these estimates to predict unobserved held-out responses in the test set.

\paragraph{Results:}
The means and standard deviations of the prediction accuracy and prediction likelihood metrics covering multiple cross-validation trials are shown in Tables~2 and 3. We see that SPARFA-Trace achieves comparable prediction performance to SPARFA-M on both datasets, although the datasets are treated as if they do not have time-varying effects. We emphasize that, in addition to providing competitive prediction performance, SPARFA-Trace is capable of (i) tracing learner concept knowledge evolution over time and (ii) analyzing learning resource and question qualities and their content organization. This extracted information is very important, as it allow a PLS to provide timely feedback to learners about their strengths and weaknesses, and to automatically recommend learning resources to learners for remedial studies based on their qualities and contents.

\begin{table}
\centering
\caption{Comparisons of SPARFA-Trace against SPARFA-M on predicting unobserved learner responses for Dataset~1.}
\label{tbl:sparfacompare220}
\vspace{0.3cm}
\begin{tabular}{lll}
\toprule
~&SPARFA-M& {\bf SPARFA-Trace} \\
\midrule
Prediction accuracy \quad \quad & $87.10 \pm 0.04 \%$ \quad \quad & \boldsymbol{$87.31 \pm 0.05 \%$}  \\ 
Prediction likelihood \quad \quad & $0.7274 \pm 0.0005$ \quad \quad & \boldsymbol{$0.7295 \pm 0.0007$}  \\ 
\bottomrule
\end{tabular}
\end{table}

\begin{table}
\centering
\caption{Comparisons of SPARFA-Trace against SPARFA-M on predicting unobserved learner responses for Dataset~2.}
\label{tbl:sparfacompare301}
\vspace{0.3cm}
\begin{tabular}{lll}
\toprule
~&SPARFA-M& {\bf SPARFA-Trace} \\
\midrule 

Prediction accuracy \quad \quad & $86.64 \pm 0.14 \%$ \quad \quad & \boldsymbol{$86.29 \pm 0.25 \%$}  \\ 
Prediction likelihood \quad \quad & $0.7037 \pm 0.0024$ \quad \quad & \boldsymbol{$0.7066 \pm 0.0028$}  \\ 
\bottomrule
\end{tabular}
\end{table}

\begin{figure*}[t]
\vspace{-0.0cm}
\centering
\subfigure[]{
\includegraphics[width=0.45\textwidth]{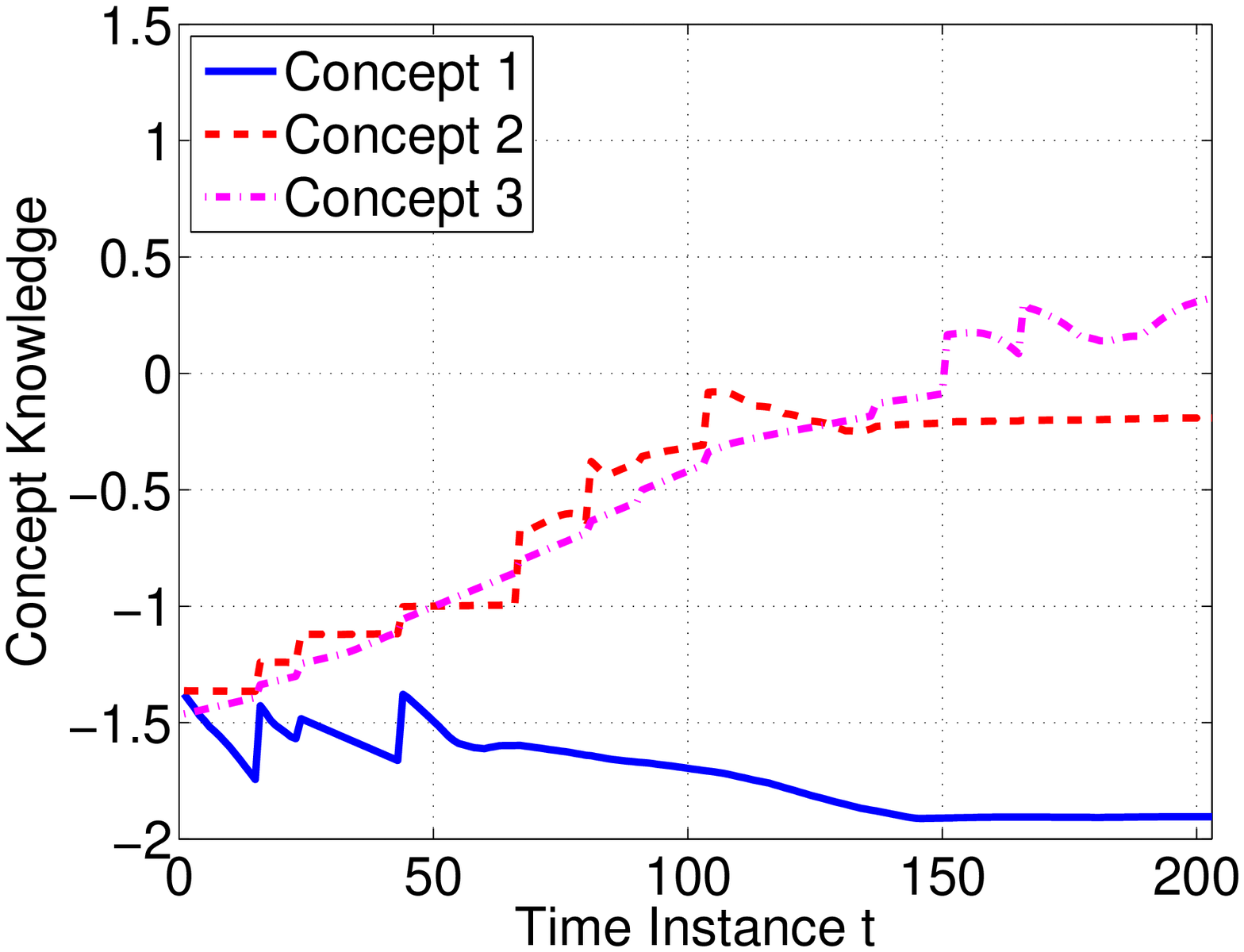}
\label{fig:cvst_1}
}
\hspace{0.6cm}
\subfigure[]{
\includegraphics[width=0.45\textwidth]{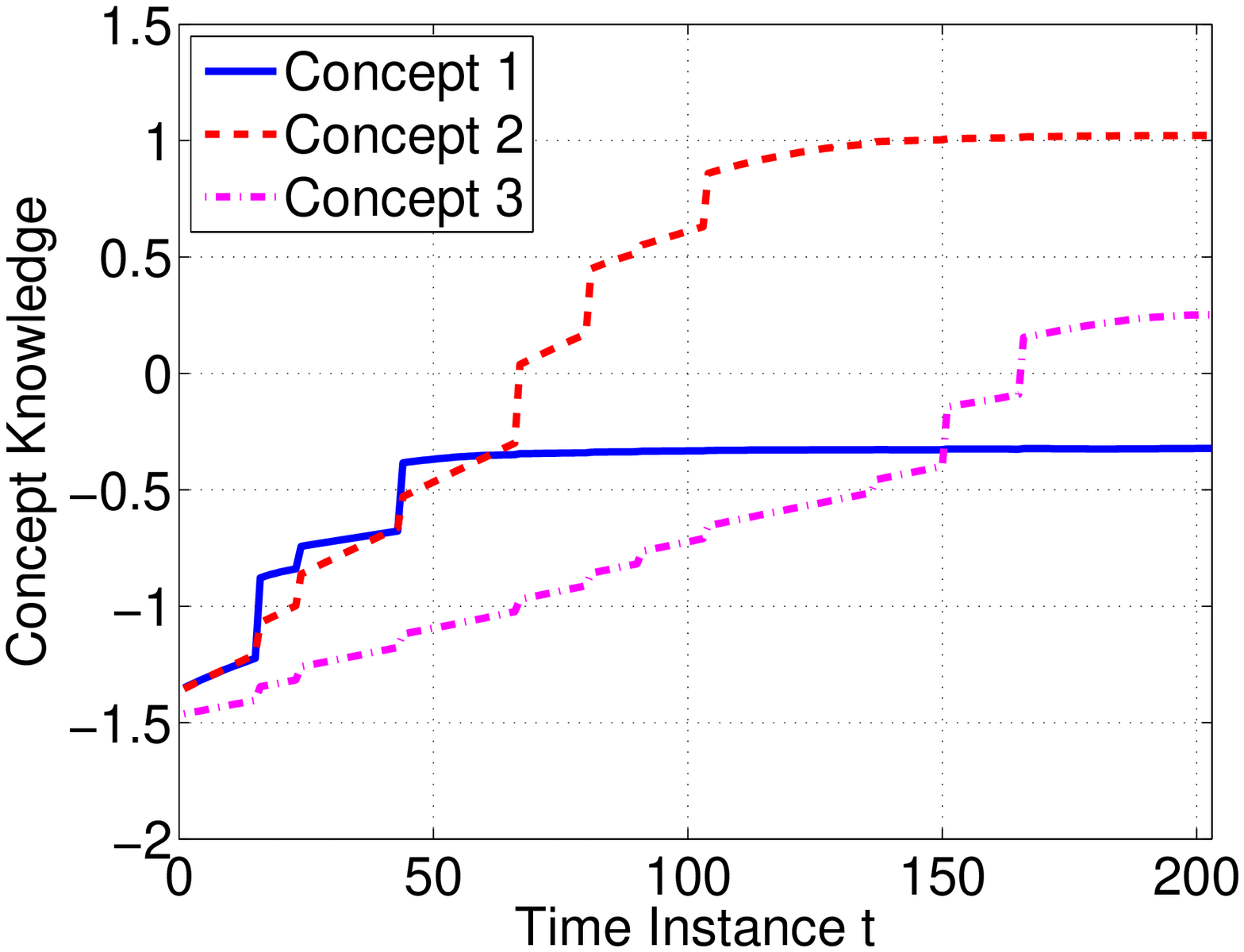}
\label{fig:cvst_m}
}\\[0.3cm]
\vspace{-0.3cm}
\caption{Estimated latent learner concept knowledge states for all time instances, for Dataset~1. (a) Learner~1's latent concept knowledge state evolution; (b) Average learner latent concept knowledge states evolution. }
\label{fig:cvst}
\vspace{-0.0cm}
\end{figure*}

\subsection{Visualizing time-varying learning and content analytics}
\label{sec:visual}

In this section, we showcase another advantage of SPARFA-Trace over existing KT and collaborative filtering methods, i.e., the visualization of both learner knowledge state evolution over time and the estimated learning resource and question quality and content organization. 

\paragraph{Visualizing learner concept knowledge state evolution:}
\fref{fig:cvst_1} shows the estimated latent learner concept knowledge states at all time instances for Learner~1 in Dataset~1. We can see that their knowledge on Concepts~2 and 3 gradually improve over time, while their knowledge on Concept~1 does not. Therefore, recommending Learner~1 remedial material on Concept~1 seems necessary, which is verified by the fact that Learner~1 often responds incorrectly on questions covering Concept~1 towards the end of the course.

\fref{fig:cvst_m} shows the average learner concept knowledge states over the entire class at all time instances for Dataset~1. Since Concept~1 is the basic concept that is covered in the early stages of the course, we can see that its mean knowledge among all learners increases in early stages of the course and then remain constant afterwards. In contrast, Concept~3 is the most advanced concept covered near the end of the course, and the improvement in which is not obvious until very late stages of the course.
Hence, SPARFA-Trace enables a PLS to provide timely feedback to individual learners on the their concept knowledge at all times, which reveals the learning progress of the learners. 
SPARFA-Trace also informs instructors on the trend of concept knowledge state evolution of the entire class, in order to help them make timely adjustments to their course plans. 

\begin{figure*}[t]
\centering
\hspace{-0cm}
\subfigure[]{
\includegraphics[width=0.4\textwidth]{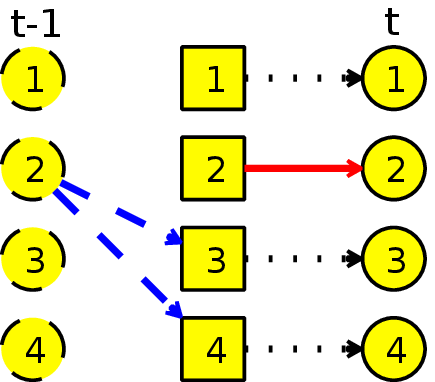}
\label{fig:Dex_3}
}
\hspace{2cm}
\subfigure[]{
\includegraphics[width=0.4\textwidth]{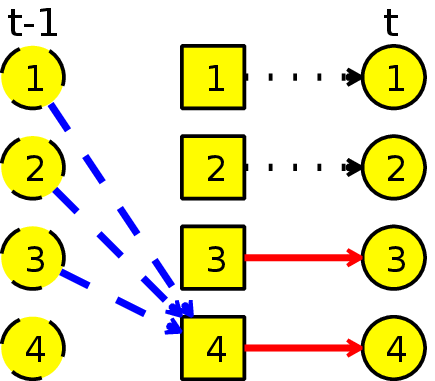}
\label{fig:Dex_9}
}\\[0.3cm]
\vspace{-0.3cm}
\caption{Visualized learner knowledge state transition effect of two distinct learning resources for Dataset~2. (a) Learner knowledge state transition effect for Learning resource~3; (b) Learner knowledge state transition effect for Learning resource~9.}
\label{fig:Dex}
\end{figure*}

\paragraph{Visualizing learning resource quality and content:}
\fref{fig:Dex_3} and \fref{fig:Dex_9} show the quality and content organization of learning resources~3 and 9 for Dataset~2. These figures visualize the leaners' concept knowledge state transitions induced by interacting with learning resources~3 and 9. 
Circular nodes represent concepts; the leftmost set of dashed nodes represent the concept knowledge state vector $\vecc^{(t-1)}$, which are the learners' concept knowledge states before interacting with these learning resources, and the rightmost set of solid nodes represent the concept knowledge state vector $\vecc^{(t)}$, which are the learners' concept knowledge states after interacting with these learning resources. 
Arrows represent the the learner concept knowledge state transition matrix $\bD_m$, the intrinsic quality vector of the learning resource $\vecd_m$, and their transformation effects on learners' concept knowledge states. 
Black, dotted arrows represent unchanged learner concept knowledge states; these arrows correspond to zero entries in $\bD_m$ and $\vecd_m$. 
Red, solid arrows represent the intrinsic  knowledge gain of some concepts, characterized by large, positive entries in~$\vecd_m$.
Blue, dashed arrows represent the change in knowledge of advanced concepts due to their pre-requisite concepts, characterized by non-zero entries in $\bD_m$: High knowledge level on pre-requisite concepts can result in improved understanding and an increase on knowledge of advanced concepts, while low knowledge level on these pre-requisite concepts can result in confusion and a decrease on knowledge of advanced concepts. 

As shown in \fref{fig:Dex_3}, Learning resource~3 is used in early stage of the course, and we can see that this learning resource gives the learners' a positive  knowledge gain of Concept~2, while also helping on more advanced Concepts~3 and 4. 
As shown in \fref{fig:Dex_9}, Learning resource~9 is used in later stage of the course, and we can see that it uses the learners' knowledge on all previous concepts to improve their knowledge on Concept~4, while also providing a positive knowledge gain on Concepts~3 and 4. 

By analyzing the content organization of learning resources and their effects on learner concept knowledge state transitions, SPARFA-Trace enables a PLS to automatically recommend corresponding learning resources to learners based on their strengths and weaknesses. The estimated learning resource quality information also helps course instructors to distinguish between effective learning resources, and poorly-designed, off-topic, or misleading learning resources, thus helping them to manage these learning resources more easily.

\begin{figure*}[tp]
\centering
\subfigure[]{
\includegraphics[width=1\columnwidth]{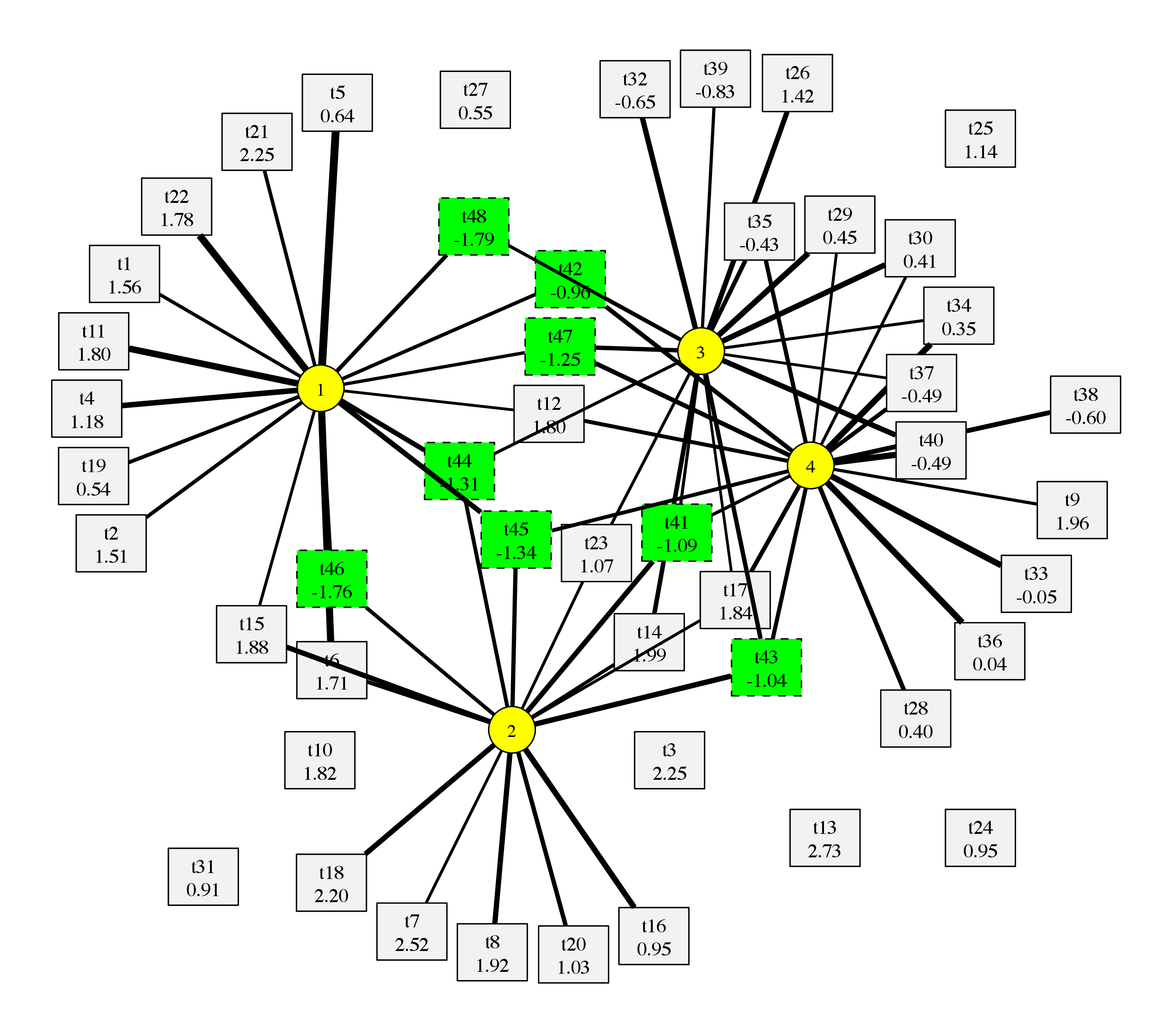}
\label{fig:Wex_a}
}\\[0.3cm]
\subfigure[]{
\scalebox{1}{%
\begin{tabular}{ll}
\toprule
\bf Concept 1 & \bf Concept 2 \\
\midrule
Laplace transform and filters & Sampling and reconstruction   \\
\midrule
\bf Concept 3 & \bf Concept 4 \\
\midrule
Fourier series and Fourier transform  & Signals and systems basics \\
\bottomrule \\
\end{tabular}}
\label{fig:Wex_b}
}
\vspace{-0.2cm}
\caption{ \subref{fig:Wex_a} Question--concept association graph and concept labels for Dataset~2. (a) Question--concept association graph. Note that for the visualization to be compact, we show only $1/3$ of all questions in the dataset; (b) Label of each concept.}
\label{fig:Wex}
\end{figure*}

\paragraph{Visualizing question quality and content:}
\fref{fig:Wex} shows the question--concept association graph obtained from Dataset~2. Yellow, circle nodes represent concept nodes, while green, box nodes represent question nodes. Each question box is labeled with the time instance at which it is assigned and its estimated intrinsic difficulty. 
From the graph we can see time-evolving effects, as questions assigned in the early stages of the course cover basic concepts (Concepts~1 and 2), while questions assigned in later stages cover more advanced concepts (Concepts~3 and 4). Some questions are associated with multiple concepts, and they mostly correspond to the final exam questions (boxes with dashed boundaries) where the entire course is covered.

Thus, by estimating the intrinsic difficulty and content organization of each question, SPARFA-Trace allows a PLS to generate feedback to instructors on the underlying knowledge structure of questions, which enables them to identify ill-posed or off-topic questions (such as questions that are not associated to any concepts in \fref{fig:Wex_a}).

\section{Related Work on Knowledge Tracing for Personalized Learning}
\label{sec:rlwork}

Various machine learning algorithms have been designed for personalized learning. 
Specifically, matrix and tensor factorization approaches have been applied to analyze graded learner responses in order to extract learner ability parameters and/or question--concept relationships. 
Examples include item response theory (IRT) (\cite{lordirt,rasch,mirt1,mirt2}), and other factor analysis models 
(\cite{qmatrix,afm,sparfa}).
While these methods have shown to provide good prediction performance on unobserved learner responses, they do not take into account the temporal dynamics involved in the process of a course. 
Therefore, these approaches are only suitable to a static testing scenario, such as the graduate record examinations (GRE), standardized tests, placement exams, etc (see \cite{adtest1} for details).

A number of approaches have also been developed to analyze temporal learner response data (see, e.g., \cite{kt,ktpardos} for details). 
In particular, knowledge tracing (KT) estimates learner concept knowledge over time, given question--concept mappings and graded binary learner response data. 
Since such methods all require pre-defined question--concept mappings which are, in general, not available in practice, these methods are labor-intensive to instructors and domain experts, and are not scalable to large-scale applications such as massive online open courses (MOOCs) (see \cite{mooc1,mooc2} for an overview).

\sloppy
Recent approaches to KT without requiring question--concept mappings, described in \cite{jose} and \cite{jose1} jointly estimate both question--concept (item--skill) mappings and learner concept mastery evolution over time purely from response data. 
Their method, however, suffers from the following deficiencies: 
First, \cite{jose} models the learners' latent concept knowledge as a small number of \emph{discrete} values and the entire dynamic process for learning is modeled as a hidden Markov model (HMM). Such discrete concept knowledge states do not provide desirable interpretability when the number of discrete learner concept knowledge values is low (the authors used $3$ distinct knowledge levels in their paper). 
In contrary, the proposed SPARFA-Trace framework models learner latent concept knowledge states as continuous random variables, providing finer knowledge representations.
Second, \cite{jose} does not handle questions that involve multiple concepts. 
In contrary, the proposed SPARFA-Trace framework directly takes into account questions involving multiple concepts in the probabilistic model.
Third, \cite{jose} introduced a Gibbs sampler approach to infer all parameters; such an approach is known to be computationally intensive and, hence, not scalable to large datasets, such as data obtained in a MOOC. 
In contrary, the proposed SPARFA-Trace framework uses a computationally efficient EM approach, which is capable of scaling to personalized learning applications at MOOC scale.
\fussy

\section{Conclusions}
\label{sec:conclusions}

We have proposed SPARFA-Trace, a novel, message passing-based approximate Kalman filtering  approach for time-varying learning and content analytics. 
The proposed method jointly traces latent learner concept knowledge and simultaneously estimates the quality and content organization of the corresponding learning resources (such as textbook sections or lecture videos), and the questions in assessment sets. 
In order to estimate latent learner concept knowledge states at each time instance from observed binary-valued graded learner responses, we have introduced an approximate Kalman filtering framework, given all learner concept knowledge state transition parameters of learning resources and the question-dependent parameters. 
In order to estimate these parameters, we have introduced novel convex optimization-based algorithms that estimate all the learner concept knowledge state transition parameters of learning resources and question--concept associations and their intrinsic difficulties. 
The proposed approach applied to real-world educational datasets has shown its capability of accurately predicting unobserved learner responses, while obtaining interpretable estimates of all learner concept knowledge state transition parameters and question--concept associations.

A PLS can benefit from the information extracted by the SPARFA-Trace framework in a number of ways.
Being able to trace learners' concept knowledge enables a PLS to make timely feedback to learners on their strengths and weaknesses. Meanwhile, this information will also enable adaptivity in designing personalized learning pathways in real time, as instructors can recommend different actions for different learners to take, based on their individual concept knowledge states. 
Furthermore, the estimated content-dependent parameters provide rich information on the knowledge structure and quality of learning resources. This capacity is crucial for a PLS to automatically suggest learning resources to learners for remedial studies. 
Together with the question parameters estimated, a PLS would be able to operate in a \emph{hands-off} manner, requiring only minimal human input and intervention; this paves the way of applying SPARFA-Trace to MOOC-scale education scenarios, where the massive amount of data prevents any manual intervention.

We finally note that a number of improvements/extensions to SPARFA-Trace could be made. 
For example, more accurate message-passing schemes like expectation propagation (\cite{qi}) could be applied to improve the performance and accuracy of SPARFA-Trace. 
More sophisticated non-affine learner concept knowledge state transition models can also be applied, in contrast to the affine model proposed in \fref{sec:modeltrans}. 
In order to provide better interpretation to the estimated learner concept knowledge state transition and question parameters, tagging and question text information can be coupled with SPARFA-Trace (see \cite{sparfatag,sparfatop} for corresponding extensions to SPARFA that mine question tags and question text information).
It is worth mentioning that SPARFA-Trace has potential to be applied to a wide range of other datasets, including (but not necessarily limited to) the analysis of temporal evolution in legislative voting data (\cite{larryvote}), and the study of temporal effects in general collaborative filtering settings (\cite{larrycf}). The extension of SPARFA-Trace to such applications is part of an on-going work.

\section{Appendix}
\label{sec:appendix}
We derive the closed-form moment matching expressions for the approximate Kalman filtering approach detailed in \fref{sec:akf}.
The following derivation can be seen as a multi-variate counterpart of the approach in \cite[Sec.~3.9]{gpml}.

We start by associating the $K$-dimensional latent variable vector $\vecc$ with a Gaussian prior
$p(\vecc) = \mathcal{N}(\vecc \!\mid\! \vecm, \bV)$,
where $\vecm$ and $\bV$ are the prior's mean and covariance matrix, respectively. The observation likelihood takes the form
$p(y\!\mid\! \vecc) = \Phi \left( \left(2y - 1\right) \left(\vecw^T \vecc - \mu \right) \right)$.
For simplicity of exposition, we will write $\widetilde{\vecw} = \left(2y - 1\right) \vecw$ and $\widetilde{\mu} = \left(2y - 1\right) \mu$ in the following derivations.
According to Bayes rule, the posterior distribution of $\vecc$ given the observation $y$ can be written as:
\begin{align*}
p(\vecc \!\mid\! y) = \frac{p(\vecc) p(y \!\mid\! \vecc)}{p(y)} = \frac{p(\vecc) p(y \!\mid\! \vecc)}{\int p(y \!\mid\! \vecc) p(\vecc) \mathrm{d}c}. 
\end{align*}
In order to approximate this posterior distribution of $\vecc$, we start by evaluating its denominator $p(y)$.
\begin{align}
\notag p(y) & = \int p(y \!\mid\! \vecc) p(\vecc) \mathrm{d}\vecc \\ 
\notag & = \int \Phi \left(\widetilde{\vecw}^T \vecc - \widetilde{\mu} \right) \mathcal{N} \left(\vecc \!\mid\! \vecm, \bV \right) \mathrm{d}\vecc\\
\notag & = \int \int_{-\infty}^{\infty} \mathcal{N} \left(t \!\mid\! 0,1 \right) \mathrm{d}t \; \mathcal{N} \left(\vecc \!\mid\! \vecm, \bV \right) \mathrm{d}\vecc\\
\notag & = \frac{1}{\sqrt{2\pi}}\frac{1}{\sqrt{(2\pi)^K |\bV|}} \int \int_{-\infty}^{\widetilde{\vecw}^T \vecc - \widetilde{\mu}} e^{-t^2/2} \mathrm{d}t \, e^{-\frac{(\vecc-\vecm)^T \bV^{-1} (\vecc-\vecm)}{2}} \mathrm{d}\vecc.
\end{align}
Now, substituting the variable $\vecc$ with $\vecc + \vecm$ and then, $t$ with $t - \widetilde{\vecw}^T \vecc$, we have
\begin{align}
\notag p(y) & = \frac{1}{\sqrt{2\pi}}\frac{1}{\sqrt{(2\pi)^K |\bV|}} \int \int_{-\infty}^{\widetilde{\vecw}^T \vecm - \widetilde{\mu}} e^{-\frac{(t-\widetilde{\vecw}^T \vecc)^2}{2}} \mathrm{d}t \, e^{-\frac{\vecc^T \bV^{-1} \vecc}{2}} \mathrm{d}\vecc\\
\notag & = \frac{1}{\sqrt{(2\pi)^{K+1} |\bV|}} \int_{-\infty}^{\widetilde{\vecw}^T \vecm - \widetilde{\mu}} \int e^{-\frac{(t-\widetilde{\vecw}^T \vecc)^2 + \vecc^T \bV^{-1} \vecc}{2}} \mathrm{d}\vecc \mathrm{d}t \\
\notag & = \int_{-\infty}^{\widetilde{\vecw}^T \vecm - \widetilde{\mu}} \int \mathcal{N} \left( \left[ \begin{array}{c} t \\ \vecc \end{array} \right] \!\mid\! \boldsymbol{0}, \left[ \begin{array}{cc} 1 & -\widetilde{\vecw}^T \\ -\widetilde{\vecw} & \widetilde{\vecw} \widetilde{\vecw}^T + \bV^{-1} \end{array} \right]^{-1} \right) \mathrm{d}\vecc\mathrm{d}t \\
 & = \int_{-\infty}^{\widetilde{\vecw}^T \vecm - \widetilde{\mu}} \mathcal{N} \left( t \!\mid\! 0, 1+\widetilde{\vecw}^T \bV \widetilde{\vecw} \right) \mathrm{d}t 
 = \Phi \left( \frac{\widetilde{\vecw}^T \vecm - \widetilde{\mu}}{\sqrt{1 + \widetilde{\vecw}^T \bV \widetilde{\vecw}}} \right). \label{eq:py}
\end{align}
\fussy
In the last two steps of this derivation, we have used the Woodbury matrix identity (\cite{hornjohnson}) and marginal Gaussian properties (\cite{gpml}). 
Since the posterior distribution is not Gaussian and prohibits the message passing procedure described in \fref{sec:kf}, our goal is to approximate it with a Gaussian distribution $q(\vecc) = \mathcal{N} (\vecc \!\mid\! \widehat{\vecm}, \widehat{\bV})$ so that the message passing procedure is tractable. 
As shown in \cite{gpml}, the specific values for $\widehat{\vecm}$ and $\widehat{\bV}$ that minimizes the Kullback-Leibler (KL) divergence between $q(\vecc)$ and $p(\vecc \!\mid\! y)$ are the first and second moments of the posterior $p(\vecc \!\mid\! y)$.
\sloppy

Next, we evaluate the first and second moments of the posterior distribution 
\begin{align*}
p(\vecc \!\mid\! y) = p(y)^{-1} \Phi \left( \widetilde{\vecw}^T \vecc - \widetilde{\mu} \right) \mathcal{N} \left( \vecc \!\mid\! \vecm, \bV\right). 
\end{align*}
where $p(y)$ is given by \fref{eq:py}. From \fref{eq:py} we can write
\begin{align}
\label{eq:pform}
\Phi \left( \frac{\widetilde{\vecw}^T \vecm - \widetilde{\mu}}{\sqrt{1 + \widetilde{\vecw}^T \bV \widetilde{\vecw}}} \right) = \int \Phi \left(\widetilde{\vecw}^T \vecc - \widetilde{\mu} \right) \mathcal{N} \left(\vecc \!\mid\! \vecm, \bV \right) \mathrm{d}\vecc.
\end{align}
Taking the derivative with respect to $\vecm$ of both sides of \fref{eq:pform} yields
\begin{align*}
 \mathcal{N} \left( \frac{\widetilde{\vecw}^T \vecm - \widetilde{\mu}}{\sqrt{1 + \widetilde{\vecw}^T \bV \widetilde{\vecw}}} \right) \frac{\widetilde{\vecw}}{\sqrt{1 + \widetilde{\vecw}^T \bV \widetilde{\vecw}}} = \int \bV^{-1} \left( \vecc - \vecm \right) \Phi \left(\widetilde{\vecw}^T \vecc - \widetilde{\mu} \right) \mathcal{N} \left(\vecc \!\mid\! \vecm, \bV \right) \mathrm{d}\vecc.
\end{align*}
Let $z = \frac{\widetilde{\vecw}^T \vecm - \widetilde{\mu}}{\sqrt{1 + \widetilde{\vecw}^T \bV \widetilde{\vecw}}}$, we have
\begin{align*}
 \mathcal{N}(z)\, \frac{\widetilde{\vecw}}{\sqrt{1 + \widetilde{\vecw}^T \bV \widetilde{\vecw}}} = \bV^{-1} \int \vecc \; \Phi\left(\widetilde{\vecw}^T \vecc - \widetilde{\mu} \right) \mathcal{N} \left(\vecc \!\mid\! \vecm, \bV \right) \mathrm{d}\vecc - \bV^{-1} \vecm \Phi(z).
\end{align*}
Thus, the mean of the posterior distribution of $\vecc$ is given by:
\begin{align}
\label{eq:mean}
\notag \mathbb{E}_{p(\vecc \mid y)}[\vecc] & = \int \vecc \, p(\vecc \!\mid\! y) \mathrm{d}\vecc \\
\notag & = \int \vecc \frac{\Phi\left(\widetilde{\vecw}^T \vecc - \widetilde{\mu} \right) \mathcal{N} \left(\vecc \!\mid\! \vecm, \bV \right)}{p(y)} \mathrm{d}\vecc\\
& = \vecm + \frac{\bV \widetilde{\vecw}}{\sqrt{1 + \widetilde{\vecw}^T \bV \widetilde{\vecw}}} \frac{\mathcal{N}(z)}{\Phi(z)}.
\end{align}
Similarly, taking the derivative with respect to $\vecm$ twice of both sides of \fref{eq:pform} yields
\begin{align*}
- z \mathcal{N}(z) \frac{\widetilde{\vecw} \widetilde{\vecw}^T}{1 + \widetilde{\vecw}^T \bV \widetilde{\vecw}} & = - \bV^{-1} \int \Phi\left(\widetilde{\vecw}^T \vecc - \widetilde{\mu} \right) \mathcal{N} \left(\vecc \!\mid\! \vecm, \bV \right) \mathrm{d}\vecc\\
& \quad + \bV^{-1} \left( \int \left(\vecc-\vecm \right) \left(\vecc-\vecm \right)^T \Phi\left(\widetilde{\vecw}^T \vecc - \widetilde{\mu} \right) \mathcal{N} \left(\vecc \mid \vecm, \bV \right) \mathrm{d}\vecc \right) \bV^{-1} \\
& = - \bV^{-1} \Phi(z) + \bV^{-1} \mathbb{E}_{p(\vecc \mid y)}[\vecc \vecc^T] \bV^{-1} \Phi(z) \\
& \quad \quad - \bV^{-1} \left(\mathbb{E}_{p(\vecc \mid y)}[\vecc] \vecm^T + \vecm \mathbb{E}_{p(\vecc \mid y)}[\vecc]^T \right) \bV^{-1} \Phi(z) \\
& \quad \quad + \bV^{-1} \vecm \vecm^T \bV^{-1} \Phi(z),
\end{align*}
where we implicitly used the fact that the covariance matrix $\bV$ is symmetric. Therefore, we have 
\begin{align*}
\mathbb{E}_{p(\vecc \mid y)}[\vecc \vecc^T] = \bV + \vecm \vecm^T +  \left(\mathbb{E}_{p(\vecc \mid y)}[\vecc] \vecm^T + \vecm \mathbb{E}_{p(\vecc \mid y)}[\vecc]^T \right) - z \frac{\mathcal{N}(z)}{\Phi(z)} \frac{\bV \widetilde{\vecw} \widetilde{\vecw}^T \bV}{1 + \widetilde{\vecw}^T \bV \widetilde{\vecw}}.
\end{align*}
Thus, the covariance of the posterior distribution is given by
\begin{align}
\label{eq:covariance}
\notag & \mathbb{E}_{p(\vecc \mid y)}[\left(\vecc- \mathbb{E}_{p(\vecc \mid y)}[\vecc]\right) \left(\vecc- \mathbb{E}_{p(\vecc \mid y)}[\vecc]\right)^T] \\
\notag & = \mathbb{E}_{p(\vecc \mid y)}[\vecc \vecc^T] - \mathbb{E}_{p(\vecc \mid y)}[\vecc] \mathbb{E}_{p(\vecc \mid y)}[\vecc]^T \\
\notag & = \bV + \vecm \vecm^T +  \left(\mathbb{E}_{p(\vecc \mid y)}[\vecc] \vecm^T + \vecm \mathbb{E}_{p(\vecc \mid y)}[\vecc]^T \right) - z \frac{\mathcal{N}(z)}{\Phi(z)} \frac{\bV \widetilde{\vecw} \widetilde{\vecw}^T \bV}{1 + \widetilde{\vecw}^T \bV \widetilde{\vecw}} - \mathbb{E}_{p(\vecc \mid y)}[\vecc] \mathbb{E}_{p(\vecc \mid y)}[\vecc]^T \\
& = \bV - \frac{\mathcal{N}(z)}{\Phi(z)} \left(z + \frac{\mathcal{N}(z)}{\Phi(z)} \right) \frac{\bV \widetilde{\vecw} \widetilde{\vecw}^T \bV}{1 + \widetilde{\vecw}^T \bV \widetilde{\vecw}},
\end{align}
where in the last step we have used \fref{eq:mean} to simplify the expression.

Thus, given the prior distribution $p(\vecc) = \mathcal{N}(\vecc \!\mid\! \vecm, \bV)$ and the observation likelihood $p(y\!\mid\! \vecc) = \Phi \left( \left(2y - 1\right) \left(\vecw^T \vecc - \mu \right) \right)$, we can approximate the posterior distribution $p(\vecc \!\mid\! y) \approx q(\vecc) = \mathcal{N} (\vecc \!\mid\! \widehat{\vecm}, \widehat{\bV})$, with $\widehat{\vecm}$ and $\widehat{\bV}$ as in \fref{eq:mean} and \fref{eq:covariance}, respectively.

\section*{Acknowledgments}

Thanks to Joe Cavallaro, Kim Davenport and JP Slavinsky for providing the OpenStax Tutor (OST) data, and Andrew Waters and Ryan Ning for helpful discussions. 
This work was supported by the National Science Foundation under Cyberlearning grant IIS-1124535, the Air Force Office of Scientific Research under grant FA9550-09-1-0432, and the Google Faculty Research Award program.
Visit our website \url{www.sparfa.com}, where you can learn more about the SPARFA project and purchase SPARFA t-shirts and other merchandise.

\bibliography{sparfaclustbib.bib}

\end{document}